\documentclass{article} % For LaTeX2e
\usepackage{iclr2027_conference,times}

\usepackage{amsmath,amsfonts,bm}

\def\eqref#1{equation~\ref{#1}}
\def\1{\bm{1}}

\DeclareMathAlphabet{\mathsfit}{\encodingdefault}{\sfdefault}{m}{sl}
\SetMathAlphabet{\mathsfit}{bold}{\encodingdefault}{\sfdefault}{bx}{n}

\usepackage{hyperref}
\usepackage{url}
\usepackage{graphicx}
\usepackage[noabbrev,capitalise]{cleveref}
\usepackage{multirow}
\usepackage{nicefrac}       
\usepackage{microtype}      
\usepackage{xcolor}         
\usepackage{gensymb}         
\usepackage{amsmath}
\usepackage{booktabs}
\usepackage{dsfont}

\title{RainAtlas: A Multi-Continental Dataset\\ for Precipitation Downscaling}

\author{
  \textbf{Pierre-Louis Lemaire} \textsuperscript{\textmd{1,}}\thanks{Work done during an internship at Mila. Corresponding author: \href{mailto:pierre-louis.lemaire@uni-tuebingen.de}{pierre-louis.lemaire@uni-tuebingen.de}}\, , \quad
  \textbf{Luca Schmidt} \textsuperscript{\textmd{2}}, \quad
  \textbf{Wietze Suijker} \textsuperscript{\textmd{3,4}}, \\
  \textbf{Alex Hernandez-Garcia} \textsuperscript{\textmd{3,5}}, \quad
  \textbf{David Rolnick} \textsuperscript{\textmd{3,6}} \\[0.8ex]
  \textsuperscript{1} connAIx Research School, Tübingen AI Center, University of Tübingen \\
  \textsuperscript{2} Cluster of Excellence Machine Learning, University of Tübingen \\
  \textsuperscript{3} Mila - Quebec AI Institute \quad
  \textsuperscript{4} IVADO \quad
  \textsuperscript{5} Université de Montréal \quad
  \textsuperscript{6} McGill University \\[0.5ex]
}

\hypersetup{
    colorlinks=true,
    allcolors=black,
    urlcolor=blue,      % color of external hyperlinks (\href, \url)
}

\iclrfinalcopy % Uncomment for camera-ready version, but NOT for submission.
\begin{document}

\maketitle

\begin{abstract}
Extreme rainfall events are increasing in intensity and frequency as climate change accelerates. While kilometer-scale precipitation forecasts are critical for supporting local decision-making, the limited availability of high-resolution precipitation observations hinders their accuracy, especially in under-resourced regions. Machine learning  models are widely used to downscale precipitation data to km-scale, but their application to unseen geographies presents challenges. First, processing raw high-resolution precipitation datasets across regions requires significant engineering and domain expertise. Second, generalization across regions remains difficult. To help overcome these barriers, we release \texttt{RainAtlas}, a large-scale, ML-ready and multi-continental dataset for precipitation downscaling. Covering three continents, \texttt{RainAtlas} harmonizes heterogeneous hourly km-scale observations to a common 2-km grid. Each regional partition contains around 210,000 aligned low- and high-resolution precipitation pairs, respectively from ERA5 reanalysis and direct observations. We benchmark state-of-the-art ML-based downscaling models across \texttt{RainAtlas} using a wide range of metrics. Our evaluation reveals substantial variance in out-of-domain generalization depending on the training regions. This underscores the need for cross-regional, multi-source km-scale evaluation, establishing \texttt{RainAtlas} as a well-positioned benchmark for precipitation downscaling research.
\end{abstract}

\section{Introduction}
\label{sec:introduction}

% Motivation for precipitation downscaling, especially at high-resolution and in region without radar networks

\begin{figure}
  \centering
  \label{fig:radar}
  \includegraphics[width=0.9\textwidth]{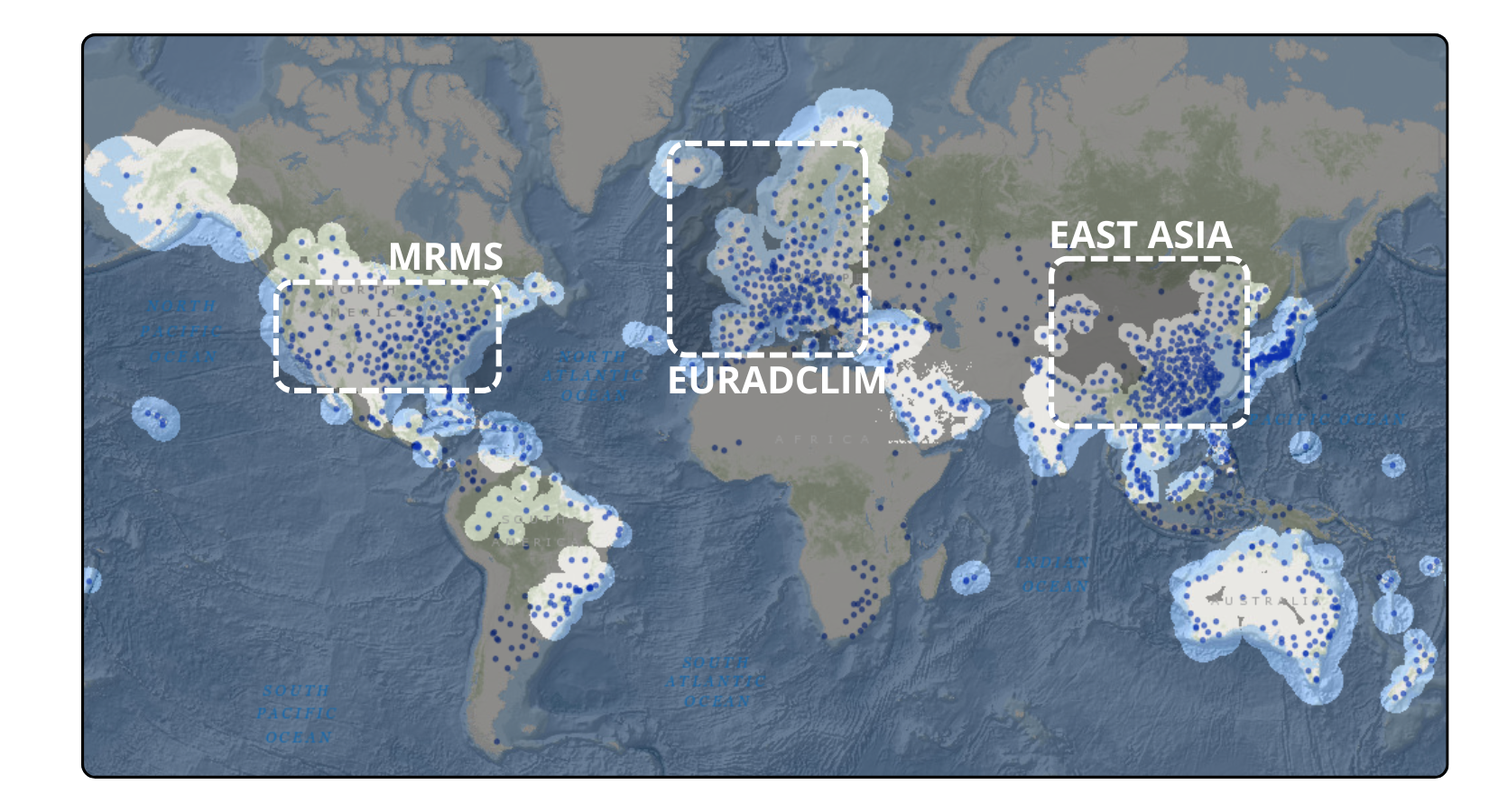}
  \caption{Global distribution of ground-based radar stations \textbf{(blue dots)} and their available coverage \textbf{(white transparent circles)}. Some regions have excellent coverage (USA, Europe, East-Asia), while other densely populated regions have little to no precipitation observations available (Africa, South America, Central Asia). \texttt{RainAtlas} consists of the three multi-source precipitation products highlighted in the figure, covering three Northern Hemisphere regions.} % TO DO: produce citation
\end{figure}

Extreme precipitation is set to increase in both intensity and frequency as a consequence of climate change \citep{cmip6-intensitification}, causing severe socio-economic consequences \citep{flashfloods, liang2022-economy}. Accurately capturing extreme precipitation events requires kilometer-scale resolution to resolve fine-scale processes such as convection and local orographic interactions \citep{convection-permitting-models}. In practice, many users leverage coarse reanalysis data products, such as ERA5, which reconstruct past weather by blending historical observations with physical models, but these severely underestimate extreme rainfall \citep{era-underestimation-extremes}, and satellite-based data products still show substantial bias \citep{hooker2026extreme}. Reliable and accurate km-scale observations remain globally scarce and unequally distributed: less than 15~\% of the world has access to accurate rain-gauge data  \citet{su2026precipitation}, while ground-based radars suffer from continent-scale coverage gaps, as shown in Figure \ref{fig:radar}. Standard physics-based climate models could theoretically bridge this gap, but their prohibitive computational cost limits widespread application \citep{neumann_assessing_2019}.

Using machine learning (ML) to \emph{downscale} precipitation data -- that is, to enhance the resolution of coarse variables (spatially, temporally, or both) -- has the potential to offer an efficient alternative, leveraging advances such as score-based models trained on historical observations to enhance spatial resolution \citep{rampal-mldownscaling-review}.
% temporary figure to show radar coverage and selected regions
Nevertheless, we identify two primary limitations hindering the practical application of ML for precipitation downscaling: (1) data processing for ML compatibility, and (2) limited geographical generalization. First, rain gauges and radars are subject to systemic bias that need to be mitigated with hydrological merging methods \citep{mckee2016-radargauge}. Additionally, radar observations require expert knowledge to be processed, differ in quality, spatial and temporal resolution, projection, format, and quality-control procedures \citep{heistermann2013open, thorndahl2017weather}. These differences create major interoperability barriers and require substantial preprocessing, including re-gridding, temporal aggregation, subsampling, and harmonization \citep{heistermann2013open}. When precipitation archives are publicly available, transforming raw multi-terabyte datasets into ML-ready datasets demands substantial computational resources and domain expertise. Second, the resulting scarcity of high-quality and ML-ready high-resolution data makes geographical generalization a challenge for ML-based downscaling models, which are rarely trained nor evaluated on multiple regions. Most approaches do not explicitly model underlying physics, implicitly assuming unchanged statistical relationships between low- and high-resolution precipitation across geographical regions. This assumption is frequently violated in practice, as precipitation patterns and extremes are governed by processes that vary substantially with local orography and regional climate phenomena. In fact, previous works report significant drops in performance when evaluating models on out-of-domain \citep{rainshift}, greatly limiting their potential for downstream applications.

These challenges highlight the need for harmonized, ML-ready and multi-region datasets to support the training and evaluation of km-scale precipitation downscaling models, encourage systematic benchmarking, and drive research efforts on geographical generalization. In response, we introduce \texttt{RainAtlas} \footnote{ All partitions of \texttt{RainAtlas} are made publicly available \href{https://huggingface.co/datasets/anonymous-cf4h776ez/rainatlas-iclr-submission}{on Hugging-Face.}} \footnote{ We release our code on an \href{https://anonymous.4open.science/r/rainatlas-anon-mirror-133F/}{anonymized repository} for review.}: a multi-source dataset and benchmark for km-scale ML-based precipitation downscaling and geographical generalization. Our dataset spans the contiguous United States, Europe, and East Asia, and include nine years of hourly observations in an ML-ready format, harmonizing three heterogeneous precipitation observation datasets to $2$ km equal-area grids. Our contributions are the following: 

% (Alex): shorter alternative to the last two paragraphs
% Benchmarks with standardized datasets and evaluation protocols offer an avenue to address these limitations, by enabling systematic model comparison and supporting methodological progress in ML for Earth system sciences. Here, we introduce \texttt{RainAtlas}: a multi-source kilometer-scale dataset collection for precipitation downscaling and geographical generalization. Our collection covers the three Northern Hemisphere continents and nine years of hourly observations in an ML-ready format, unifying three heterogeneous precipitation observations datasets to a $2$ km grid. Our contributions are the following: 

\begin{itemize}
    \item We release, to our knowledge, the first harmonized, multi-continental dataset for km-scale precipitation downscaling and geographical generalization, including over 1.2M samples.
    \item We introduce a modular framework for generating ML-ready precipitation datasets with flexible spatial and temporal subsampling configurations.
    \item We benchmark ML-based downscaling models for geographical generalization.
\end{itemize}

\section{Related work}

\paragraph{ML-based precipitation downscaling}
Generative models account for the majority of recent work on ML-based precipitation downscaling. \citet{price2022} applied conditional GANs to radar observations to downscale coarse precipitation fields to km-scale. Score-based and diffusion models have also gained significant traction for km-scale precipitation downscaling \citep{yuhao2025, dai2025precipdiff, chugang2026}. In particular, \citet{mardani2025-corrdiff} proposed a two-step \textit{corrective} diffusion model that demonstrated strong skill in reconstructing high spatial frequencies and precipitation extremes. More recently, \citet{hess-consistency} trained a consistency model for unpaired km-scale downscaling, achieving similar performance as standard diffusion, at a fraction of its computational cost. Finally, \citet{keisler2026serpentflow} introduced SerpentFlow, an unpaired framework for climate downscaling using flow matching. Despite these advances, almost all exclusively focus on single geographical regions. Only a few studies evaluate models across multiple domains \citep{glawion2025global, vicensmiquel2025}, while none investigate multi-region training.

\paragraph{Benchmarks for precipitation downscaling} 
Benchmarks with standardized datasets and evaluation protocols offer an avenue to address these limitations, by enabling systematic model comparison and supporting methodological progress in ML for Earth system sciences. RainBench \citet{de2021rainbench} proposed a global, multi-modal benchmark for precipitation forecasting at a resolution around 30 km. RainNet \citet{chen2022rainnet} provided a fine-scale precipitation downscaling dataset, with a target resolution of 4 km, but is limited to the U.S. East Coast. RainShift \citet{rainshift} introduced a global benchmark for evaluating geographical transferability in precipitation downscaling. However, it considers a comparatively coarse target resolution of around 10 km, and uses satellite observations as ground truth. Therefore, geographical generalization at km-scale resolution remains largely unexplored for ML-based precipitation downscaling.

\section{\texttt{RainAtlas}: kilometer-scale precipitation downscaling across continents}
\label{sec:dataset}

\texttt{RainAtlas} is a multi-continental dataset and benchmark designed for km-scale precipitation downscaling and geographical generalization across diverse regions. It harmonizes high-resolution observations from heterogeneous sources with ERA5 reanalysis and static covariates over large continental domains, providing aligned low- and high-resolution pairs ready for ML-based downscaling. Train, validation and test datasets are subsampled from the harmonized domain-wide datasets along their spatial and temporal dimensions. Our framework is built with a modular structure to allow for the addition of new observational datasets and subsampling strategies.

% Figure~\ref{fig:pipeline} summarises the four-panel data pipeline: heterogeneous radar archives, equal-area harmonisation onto per-region $2$~km grids with ERA5 and static covariates, importance- or random-sampled $256 \times 256$ crops across five training-set variants, and the $4 \times 5$ model-family $\times$ dataset evaluation surface used in our benchmark.

% \begin{figure}
%   \centering
%   \includegraphics[width=\textwidth]{figures/pipeline_diagram.pdf}
%   \caption{\texttt{RainAtlas} data pipeline. Three native-grid radar archives (MRMS, EURADCLIM, EA) are harmonised onto per-region $2$~km equal-area grids, co-located with ERA5 ($9$ atmospheric variables at $0.25^\circ$) and static covariates ($3$ high-resolution layers), then importance- or random-sampled into $256 \times 256$ crops across five training-set variants. The right-most panel previews the $4 \times 5$ model-family $\times$ dataset evaluation surface; importance sampling lifts the $>\!50\%$ rainy-crop fraction by $16$/$5$/$18\times$ over uniform random. Each shard ships a \texttt{\_meta.json} manifest with SHA-256, row count, and byte size for integrity verification.}
%   \label{fig:pipeline}
% \end{figure}

\subsection{High-resolution datasets}

\paragraph{United States} We use the Multi-Radar/Multi-Sensor System (MRMS) precipitation monitoring system \citet{mrms}. MRMS quantitative precipitation estimation products automatically integrate data streams from a network of over $200$ ground-based radars, satellites, numerical weather prediction (NWP) models, and rain gauges. MRMS has a native resolution of $0.01\degree \times 0.01\degree$ on a regular latitude-longitude grid. Hourly observations from the most recent operational version are available from late $2020$ onward; we include data from $2021$ to $2025$.

\paragraph{Europe} We use the EURADCLIM dataset \citet{euradclim}, also referred to as EUR. EURADCLIM provides precipitation observations on a $2$ km grid, integrating observations from $138$ ground-based radars and more than 7,700 rain gauges. Around $78~\%$ of Europe is covered, with data missing only for Italy. EURADCLIM contains a decade of precipitation observations, from $2013$ to $2023$. We include a subset of this period ($2017$ to $2022$) in \texttt{RainAtlas}.

\paragraph{East Asia} We use a km-scale dataset that covers most of China and some neighboring regions, referred to as EA or EASTASIA \citep{eastasia-data}. EA integrates ground-based radar observations, IMERG satellite data, and around 2,700 rain gauge measurements using machine learning \citep{eastasia-methods}. The resulting dataset has a native $0.01\degree \times 0.01\degree$ regular latitude-longitude grid, and covers $2017$ to $2022$. 

\subsection{Reanalysis and static covariates}

% present ERA5 and static variables, list of variables and impact

ML-based precipitation downscaling aims to provide high-resolution precipitation forecasts or projections globally, especially for data-sparse regions. This requires model inputs, such as coarse forecast or reanalysis fields and auxiliary covariates, to be globally available.
%, even when initial conditions are missing or sparse. 
%Coherently, the coarse inputs and other covariates must be available globally.

As coarse inputs, we use the ERA5 atmospheric reanalysis from the European Centre for Medium-Range Weather Forecasts (ECMWF) \citep{era5}, which assimilates hourly numerical weather predictions (NWP) with observations at a global $0.25\degree \times 0.25\degree$ regular latitude-longitude resolution. Following previous work and domain knowledge \citep{rainshift, ecpoint}, we select nine variables known to have an impact on high-resolution precipitation, which we present in Table \ref{vars-table}.

As static covariates, we use the land-sea mask derived from NASA's MODIS elevation product at its native 250 m resolution \citep{lsm}. Coastal regions are known for having complex precipitation patterns, due to the frontier between oceanic and continental atmospheric processes. Elevation strongly influences local climates, so we also include the average, as well as the standard deviation of elevation from the 1 km resolution GMTED2010 dataset \citep{elevation}.

\begin{table}
  \caption{ERA5 and static input data variables.}
  \label{vars-table}
  \centering
  \begin{tabular}{lllll}
    \toprule
    Variable     & Description     & Unit & Level \\
    \midrule
    tp & Total precipitation  & mm & surface    \\
    cp & Convective precipitation & mm & surface   \\
    cape & Convective potential energy & J$\cdot$ kg$^{-1}$ & surface   \\
    sp & Surface pressure & Pa & surface \\
    tisr & Top-of-the-atmosphere incident solar radiation & J $\cdot$ m$^{-2}$ & surface \\
    tcw & Total column water & kg $\cdot$ m$^{-2}$ & total column  \\
    tclw & Total column cloud liquid water & kg $\cdot$ m$^{-2}$ & total column  \\
    u & Eastward wind velocity & m $\cdot$ s$^{-1}$ & $700$~hPa   \\
    v & Northward wind velocity & m $\cdot$ s$^{-1}$ & $700$~hPa  \\
    \midrule
    lsm & Land-sea mask & (0, 1) & surface \\
    elev$_{mean}$ & Elevation mean & m & surface \\
    elev$_{std}$ & Elevation standard deviation & m & surface \\
    \bottomrule
  \end{tabular}
\end{table}

\subsection{Re-projection to equal-area kilometer grids}
\label{sec:reprojection}

Regular latitude-longitude grids introduce distortions in grid cell area. For example, a $0.01 \degree \times 0.01 \degree$ grid cell in Oslo, Norway would cover $0.56$ km$^2$, compared to $0.94$ km$^2$ for San Diego, USA. To reduce geographical bias and better preserve translation invariance, we reproject all low- and high-resolution datasets to region-specific equal-area grids. 

MRMS, EA, and their corresponding ERA5 and static covariates are reprojected to Albers equal-area conic projections centered on their domains, which minimizes shape distortions for regions with large east-west extent. EURADCLIM already uses the Lambert azimuthal equal-area projection, commonly used for European domains. We use its projection to reproject ERA5 and the static variables over the European domain.

We resample the MRMS and EA target datasets to a common $2 \times 2$ km resolution, and all ERA5 fields to a $24 \times 24$ km resolution, using nearest-neighbour resampling. EURADCLIM is natively on a $2$ km grid and is not resampled. This results in a downscaling factor of $12$.

\subsection{Preprocessing and format}

Raw precipitation observations contain coverage heterogeneities, artifacts, and unphysical extremes that must be corrected. First, we mask out parts of the MRMS spatial domain to ensure regions containing only upsampled coarse forecasts are removed. For EURADCLIM, we identify phyiscally incoherent extremes and filter out any timestep containing values above a conservative threshold aligned the data with historical European records. We additionally remove isolated artifacts in EURADCLIM and EASTASIA. We automatically ensure complete spatial alignment between high-resolution targets and ERA5 upon reprojection and enforce temporal consistency across hourly accumulations. Finally, all pre-processed high-resolution observational targets and their corresponding ERA5 input datasets are converted into cloud-optimized \textit{Zarr} stores, chunked along the temporal dimension, and compressed without loss of information. Static covariates are stored separately in standardized NetCDF files. We provide additional details in Appendix~\ref{sec:appendix-preprocessing}.

% Each high-resolution observational dataset, along with its corresponding aligned ERA5 dataset, is converted into a common cloud-optimized format (\textit{Zarr}) following a standardized layout. Preprocessing steps are applied to ensure that only regions with sufficient observational coverage are retained where applicable, enforce spatial and temporal alignment between ERA5 and the high-resolution datasets, and remove artifacts and incoherent extremes values (see Appendix \ref{sec:appendix-preprocessing}). Static datasets are stored separately in standardized NetCDF files. 

%All datasets whose licenses permit redistribution, as well as the preprocessing code, are made publicly available (Appendix~\ref{sec:appendix-access}).

\subsection{Stochastic spatio-temporal subsampling of precipitation events}
\label{sec:subsampling}

Since the spatial domains of the source datasets included in \texttt{RainAtlas} are rather large (\textit{e.g.,} 1,900 $\times$ 2,100 grid cells for EURADCLIM), we split them into spatial crops of dimensions $256 \times 256$, each covering around $512 \times 512$ km$^2$. Potential crops are delineated using a sliding window with $32$ strides, and $32$ margins, and assigned a sampling probability. All crops with more than $75~\%$ of missing values are discarded, which removes crops outside the actual dataset's coverage, but still includes partial coverage gaps and coastal regions. When a crop is sampled, random horizontal and vertical offsets, ranging between $-32$ and $32$ grid cells, are applied to ensure exhaustive coverage of the spatial domain \citep{dgmr}.

Precipitation datasets are highly sparse, dominated by dry events where no measurable rain occurs. An analysis of randomly subsampled datasets from MRMS, EURADCLIM, and EA (see Table \ref{tab:statistics} in Appendix \ref{sec:appendix-subsampling}) shows that only $0.88~\%$, $8.89~\%$ and $0.75~\%$ of crops, respectively, contain more than $50~\%$ non-zero precipitation values. Training on a purely random distribution of spatial crops may lead to unreliable estimates of comparatively rare, heavy precipitation events, which are of great importance for downstream applications.

To mitigate this, we adopt an importance sampling strategy that increases the relative probability of selecting crops with heavier precipitation, based on the rain-rate saturation logic proposed by \citet{dgmr}. For each candidate crop $x_n$, we compute an acceptance score $S_n$ by aggregating grid-cell intensities across the spatial dimensions:

\begin{equation}
    S_n(x_n) = \min \left\{ 1, \frac{m}{C} \sum_{c=1}^{C} g(x_{n, c}) \right\} \text{\,,}
\end{equation}

where $C$ is the total number of grid cells in the crop, and $m$ is a scaling multiplier. The term $g(x_{n, c})$ represents the rain-rate saturation function:

\begin{equation}
    g(x) = 1 - \exp(-x/s) \text{\,.}
\end{equation}

Here, $s$ is a saturation constant that controls the sensitivity to high-intensity rain. The exponential form ensures that the score $S_n$ reflects the spatial extent of significant precipitation rather than being dominated by a single extreme pixel value. Once the scores of all candidate crops are computed, the final acceptance probability $p_n$ is obtained by normalizing $S_n$ by the total sum of scores across all potential crops. Throughout our experiments, we use $m=0.1$ and $s=1.0$ for training and validation, and $m=0.2$ and $s=30.0$ for test, as suggested in \citet{dgmr} for MRMS. We find that this subsampling strategy significantly increases the representation of heavy precipitation events in the training data, as shown in Table \ref{tab:statistics}, and use it for all our experiments. Future work may explore the impact of alternative sampling strategies.

\section{Evaluation}

\subsection{Temporal and geographical generalization}

% temporal split and generalization

Due to the limited availability and uneven temporal coverage of the high-resolution precipitation products, we must often work with observations datasets that do not fully overlap in time. The atmosphere is a chaotic system, with an effective predictability limit of approximately 14 days \citep{lorenz-fourteen-days}. Therefore, we assume that there is no data leakage when using recent observations for training while evaluating on past observations. We note that, to minimize overfitting to global climate patterns, ML models must be trained and evaluated on distinct time periods when assessing geographical transferability. Given the temporal coverage of \texttt{RainAtlas}'s source datasets, we use observations from $2021$ for validation, $2022$ for test, $2017$ to $2020$ for subsampling training crops from EURADCLIM and EA, and $2023$ to $2025$ for MRMS. A visual representation of the splits is proposed in Figure \ref{fig:train-val-test-splits-years} in Appendix \ref{sec:appendix-architectures}.

% geographical generalization across continents
To evaluate geographical generalization across distinct climatic regimes, we additionally perform for following train-test splits: (1) train on each single region and test on the remaining two held out regions, and (2) train on each pair of regions and test on the remaining region.

\subsection{Baseline models}
We compare several state-of-the-art ML-based downscaling approaches: a deterministic UNet and four generative models, namely an EDM-style diffusion model, \textit{corrective} diffusion, SerpentFlow and a consistency model. SerpentFlow is designed for unpaired data; we train the consistency model on paired data. All models share the same UNet backbone and are conditioned on the static covariates and the bilinearly upsampled ERA5 variables. We do not include any temporal context. As a non-ML reference, we also report the bilinearly upsampled ERA5 \textit{total precipitation}. %The downscaling methods differ in their training objective and how they formulate or decompose the downscaling task.

\paragraph{UNet}
We adopt the UNet architecture introduced in ClimateDiffuse \citep{watt2024generative}. The model is trained to predict the residual between the high-resolution target and the bilinearly upsampled ERA5 \textit{total precipitation} variable with $l_2$ loss.

\paragraph{EDM-style Diffusion}
We adopt the conditional diffusion approach from ClimateDiffuse \citep{watt2024generative}, based on EDM diffusion framework proposed by \citet{karras2022edm} that optimizes noise scheduling, network preconditioning, and 2nd-order ODE sampling for faster, more stable training and generation.

\paragraph{\textit{Corrective} Diffusion}
Inspired by CorrDiff \citep{mardani2025-corrdiff}, \textit{corrective} diffusion decomposes the downscaling task into two stages. First, a deterministic UNet predicts a conditional mean. A diffusion model then learns to generate the conditional variance of the high-resolution target. For both stages, we reuse the UNet and EDM-style diffusion model described above.
 
\paragraph{SerpentFlow}
Following \citet{keisler2026serpentflow}, we use SerpentFlow: a two-step generative framework for downscaling with unpaired data. SerpentFlow first identifies a scale below which the low- and high-resolution datasets can no longer be distinguished, using low-pass filtering and a classifier as a cut-off criterion. After replacing the domain specific high-resolution information with noise, a model is trained to recover it through a flow-matching objective. Unlike the other models, SerpentFlow is not conditioned on the ERA5 \textit{total precipitation} variable, but ERA5 fields are used to define the shared coarse domain and as part of the input at inference.

\paragraph{Consistency model}
We further include a consistency model \citep{song2023consistency, song2024improved}. The model is trained with a consistency objective that encourages predictions from different noise levels along the same probability-flow ODE trajectory to agree. This enables high-fidelity generation in a single step, bypassing the costly iterative sampling required by standard diffusion models, while still permitting multi-step sampling to trade compute for fidelity.

Further details about architectures and training algorithms are provided in Appendix \ref{sec:appendix-architectures}.

\subsection{Metrics}

Precipitation downscaling is a multi-faceted task \citep{review-precipitation-downscaling}, for which no single metric or unified score can adequately assess model performance. Depending on the downstream applications and end-user needs, different aspects of the evaluation may vary in importance. We therefore evaluate a diverse set of metrics capturing point-wise accuracy, scale-dependent skill, spectral and distributional fidelity, and probabilistic calibration. We detail mathematical formula in Appendix \ref{sec:appendix-metrics}.

\paragraph{Point-wise and probabilistic errors (MAE, CRPS and bias)} We compute the bias, the mean absolute error (MAE) and its probabilistic analogue, the continuous ranked probability score (CRPS) using an 8-member ensemble \citep{hans2000-crps}. Pointwise errors are affected by a double-penalty problem. To reduce sensitivity to the small-scale displacement errors, we additionally evaluate these metrics after spatially averaging predictions and observations by factors of 4, 8 and 12. %Since extreme precipitation events are of greater interest in practical applications and precipitation is heavy-tailed, we also compute the quantile-wise MAE, and a weighted CRPS focused on the upper tail \citep{gneiting2011-utcrps}.

\paragraph{Spatial and scale-dependent skill (FSS and variogram)} 
To circumvent the double-penalty problem and assess intense precipitation accuracy, we compute the fractions skill score (FSS) \citep{roberts2008scale} across multiple spatial neighbourhood radius and precipitation intensity thresholds. Additionally, we compute the variogram score \citep{variogram-score}, a proper scoring rule that penalizes errors in the spatial structure and pairwise dependencies of ensemble predictions.
 
\paragraph{Spectral and intensity fidelity (RAPSD and LHD)} We compute the radially averaged power spectral density (RAPSD) to evaluate whether ML models are able to reproduce the observed distribution of spatial variability across scales \citep{ruz2011-rapsd}. As for the fidelity of precipitation intensities, we analyze intensity histograms and compute their logarithmic distance (LHD) relative to observations.

\paragraph{Calibration and uncertainty (SSR and rank histograms)} Rank histograms are computed for generative models to measure if their ensemble spread capture the true variability of the observations \citep{hamill2001-rankhistograms}. We also compute the spread-skill ratio, which benchmarks internal model uncertainty against average error to further evaluate ensemble calibration. 

\section{Results}

\subsection{In-domain performance}

\begin{figure}[h]
    \centering
    \includegraphics[width=1.0\linewidth]{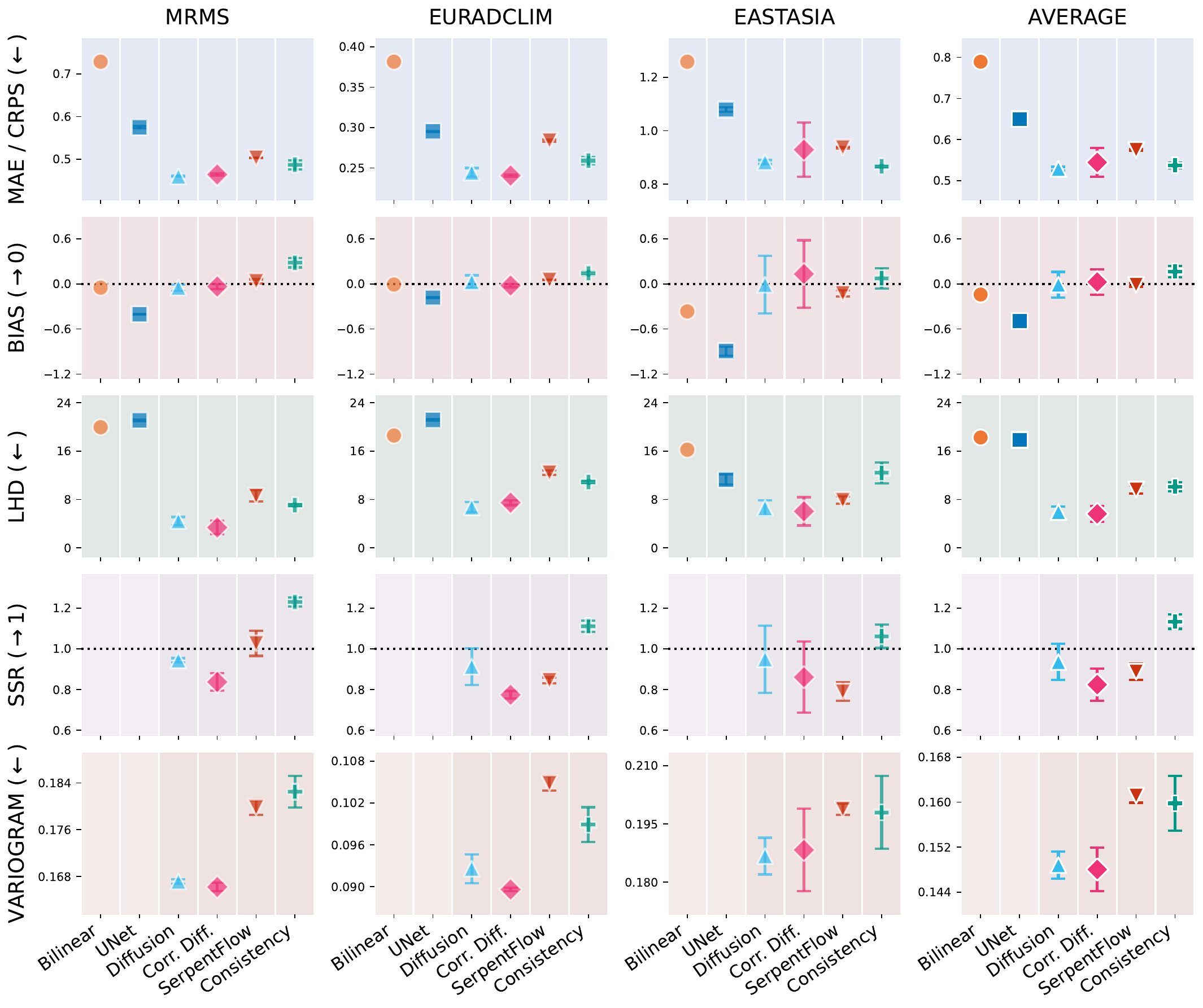}
    \caption{\textbf{Diffusion-based models outperform other baselines.} In-domain performance evaluation on importance-subsampled test sets. MAE/CRPS, and bias are computed without artificial coarsening. MAE/CRPS, and variogram rows have region-specific $y$-axis ranges to facilitate the visualization. Arrows indicate the desired direction for each metric. Numerical results are detailed in Table \ref{tab:indomain-results-importance}.}
    \label{fig:indomain-performance-viz}
\end{figure}

Figure \ref{fig:indomain-performance-viz} presents a visualization of in-domain unit scores across regions on the importance-subsampled test datasets. The numerical results are provided in Table \ref{tab:indomain-results-importance}.

Generative models outperform deterministic baselines across the evaluation metrics. Trained on the conditional mean, the UNet baseline produces excessive smoothness and fails to capture high-intensity precipitation events, as shown by its poor performance on spectral and distributional fidelity (see Figure \ref{fig:indomain-histograms-rapsd-importance}), and its low FSS on intensites above 5 \texttt{mm} (see Figure \ref{fig:fss-importance}).  Bilinear interpolation of ERA5 maintains near-zero bias, except in EA (which indicates a misalignment between ERA5 and the km-scale observational dataset), but results in a much larger MAE and similar failures as the UNet. 

Regarding generative baselines, diffusion-based models show consistently low CRPS and high intensity fidelity across the three datasets. However, we note that the \textit{corrective} diffusion model slightly under-perform on spectral fidelity compared to the diffusion-only model. This stems from the UNet's poor spectral fidelity, upon which the \textit{corrective} diffusion model is conditioned on, which creates a larger RAPSD gap to recover compared to bilinear interpolation.  Despite being trained on unpaired data, SerpentFlow exhibits a surprisingly low overall bias across datasets. Similarly, while we employ only one-step sampling, results from the consistency model yield relatively low CRPS, but high variance when predicting extreme precipitation across all spatial scales, as shown by its large relative RAPSD for each region (see Figure \ref{fig:indomain-histograms-rapsd-importance}, bottom). Additionally, both SerpentFlow and the consistency models suffer from right-tail overestimation as reflected by their poor LHD scores, which hinders their overall performance. We find all baselines be well calibrated, the diffusion model outperforming others, and the consistency model being slightly over-dispersed as a result of its higher variance. Overall, the diffusion model outperform the other generative models in-domain, scoring best or second across all metrics.

\subsection{Geographical generalization}

In this section, we choose to focus our main analysis on pointwise errors and spatial structure, since it is more challenging to improve these metrics through post-hoc bias correction and quantile mapping techniques. Additional results for the rest of the metrics are provided in Appendix \ref{sec:appendix-geographical-generalization}. 

\begin{figure}
    \centering
    \includegraphics[width=1.0\linewidth]{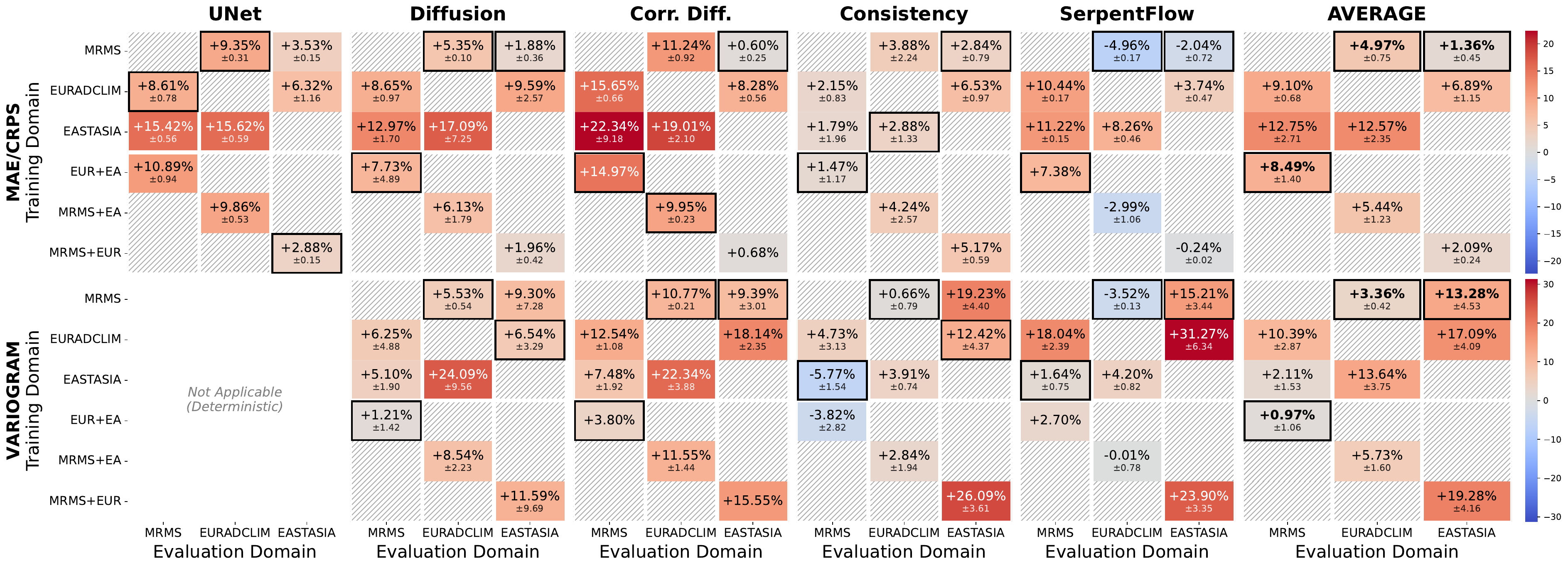}
    \caption{\textbf{Models trained on MRMS show more robust geographical generalization.} Relative MAE, CRPS and VARIOGRAM degradation ($\%$) with respect to in-domain training. Additional results are detailed in Figures \ref{fig:appendix-generalization-importance} and \ref{fig:appendix-generalization-random}.}
    \label{fig:generalization-importance}
\end{figure}

As can be seen on Figure \ref{fig:generalization-importance}, the generalization performance on unseen regions relative to in-domain results is mostly determined by the training dataset. What stands out is the generally moderate out-of-domain performance degradation when models are trained on the MRMS dataset only, with a maximum relative increase in CRPS, among generative baselines and evaluation domains, of $11.24\%$ for the \textit{corrective} diffusion model when evaluated on EUR. On average, probabilistic models trained on MRMS yield a relative CRPS degradation of $4.97\%$ on EUR, and $1.36\%$ on EA. Contrastively, the UNet shows the most important CRPS degradations across baselines, with $9.35\%$ and $3.53\%$ for EUR and EA respectively. Surprisingly, with the exception of the UNet model, combining another region with MRMS during training did not enhance geographical performance, however it almost consistently reduced degradation compared to solely training on the additional region. CRPS degradation drops from $12.57\%$ to $5.44\%$ on average when evaluated on EUR, and from from $6.89\%$ to $2.09\%$ on EA. Results for the variogram-score show a similar agreement, with performance maintaining a similar region-wise ranking. The average out-of-domain relative degradation on EUR equals to $3.36\%$ versus $13.64\%$ when trained respectively on MRMS or EA, and similarly when evaluated on EA, the models trained on MRMS show a $13.28\%$ relative degradation, compared to $17.09\%$ when trained on EUR. There two likely factors explaining these results: (1) MRMS is the most homogeneous product in terms of observational sources, with the most ground-based radars from a single unified radar network, and (2) it is the most diverse domain in terms of climatic conditions \citep{koppen-geiger-climates}. An interesting research avenue for future work would hence be investigating whether sampling from specific climate types, related to the target domains, improves geographical generalization. 

We now turn to the geographical generalization specifically on the MRMS dataset. Interestingly, we observe the opposite trend in this case: higher degradation from single-region training than with multi-region training. While reaching high increase in relative CRPS when trained solely on EUR or EA, models are able to better generalize to the MRMS domain when trained on the two other datasets combined. This observation is specifically true for the diffusion-based models, which achieve substantial improvements in variogram-score: from $6.25\%$ (EUR) and $5.10\%$ (EA) to $1.21\%$ degradation for the diffusion baseline, and from $12.54\%$ (EUR) and $7.48\%$ (EA) to $3.80\%$ degradation for \textit{corrective} diffusion. Again, we observe an important agreement on average results between CRPS and the variogram-score, showing that training on a combination of regions can significantly improve generalization performance when individual regions include limited climatic diversity, and suffer from more important observational bias: incoherent extreme values for EUR, and scarcer radar coverage for EA. 

It is worth noting that among all baselines, the consistency model shows substantially better and more stable geographical generalization, with almost all relative CRPS degradation under $5\%$ across training and evaluation configurations. We also find that SerpentFlow achieves low CRPS degradation, even improving in-domain performance in some cases. While this can be explained by its weaker in-domain CRPS, it still shows that unpaired downscaling can be a promising avenue for geographical generalization. 

Another interesting finding is that the \textit{corrective} diffusion model suffers on average from higher degradation than other generative baselines. A plausible explanation is that its input already diverges significantly, as shown by the UNet's high relative degradation on held-out domains. In fact, if we compare the results from these three baselines, we can infer that the \textit{corrective} diffusion model's degradation follows an approximate addition of the UNet's and the diffusion's degradation. 

\section{Conclusion}

We introduced \texttt{RainAtlas}, a harmonized, multi-continental dataset designed to address interoperability and generalization barriers in precipitation downscaling at km-scale.  By standardizing heterogeneous radar, satellite and rain-gauges observations onto equal area grids and pairing them with ERA5 reanalysis and static covariates, we provide a modular framework tailored for machine learning applications. Benchmarking state-of-the-art architectures reveals that while generative models capture complex spatial patterns in-domain, generalizing to unseen geographies remains challenging. We note substantial variance in generalization performance, that we suggest depends heavily on the climatic conditions diversity of training datasets. Furthermore, we observe that models relying on sequential predictions, such as \textit{corrective} diffusion, might suffer from compounded errors under geographical shift. These findings underscore the need for unified geographical benchmark in precipitation downscaling. By providing a multi-continental ML-ready benchmark, \texttt{RainAtlas} could help the Earth system science community more systematically investigate transferability dynamics, and support the development of models that produce reliable high-resolution precipitation estimates to regions lacking observational infrastructure.

\subsection*{Acknowledgments}

This work was financially supported by the Schmidt Sciences AI2050 program, the Canada CIFAR AI Chairs program and IVADO. Computational resources were provided by Mila (\href{https://www.mila.quebec}{Mila}) and the Digital Research Alliance of Canada (\href{https://www.alliancecan.ca}{Alliance Canada}). We thank Francis Pelletier for his assistance with the implementation of the data loading and processing modules, as well as Fenwick Cooper and Shruti Nath for their valuable feedback, useful suggestions, and fruitful conversations. 

\subsection*{AI use statement}

In this work, we used generative AI tools for improving the writing quality of our original drafts, as well as guidance for shortening some sections of the main text. We also used AI tools for assisting with parts of our code implementation, as well as for generating the code that produced the figures we present. We have not used generative AI tools for drafting entire sections of the paper nor for idea generation. We have reviewed all AI-assisted work. We take responsibility for the final content of this work, including text, claims or artifacts produced with the aid of generative AI.

\subsection*{Ethics statement}

\texttt{RainAtlas} is built from public observation products. MRMS is produced by NOAA and is in the public domain. EURADCLIM is released by KNMI under CC BY 4.0. The EA product is distributed by TPDC. The data contain no personal information. Downscaled precipitation from models trained on \texttt{RainAtlas} should not be used for operational warnings or risk decisions without local validation, in particular outside the three regions, since our results show that evaluation on unseen regions can degrade skill.

\subsection*{Reproducibility statement}

Dataset construction and preprocessing are described in Section~\ref{sec:dataset} and Appendix~\ref{sec:appendix-preprocessing}; the subsampling procedure in Section~\ref{sec:subsampling}; architectures, training details and hyperparameters in Appendix~\ref{sec:appendix-architectures}; and all metric definitions in Appendix~\ref{sec:appendix-metrics}. The dataset is available on Hugging Face. We release the \href{https://anonymous.4open.science/r/rainatlas-anon-mirror-133F/}{code} for preprocessing, training and evaluation, and the seeds used for training ($42$, $84$, $126$).

\bibliography{references}
\bibliographystyle{iclr2027_conference}

\appendix

\section{Preprocessing of low- and high-resolution source datasets}
\label{sec:appendix-preprocessing}

\begin{figure}[h]
  \centering
  \label{fig:mrms-mr}
  \includegraphics[width=0.9\textwidth]{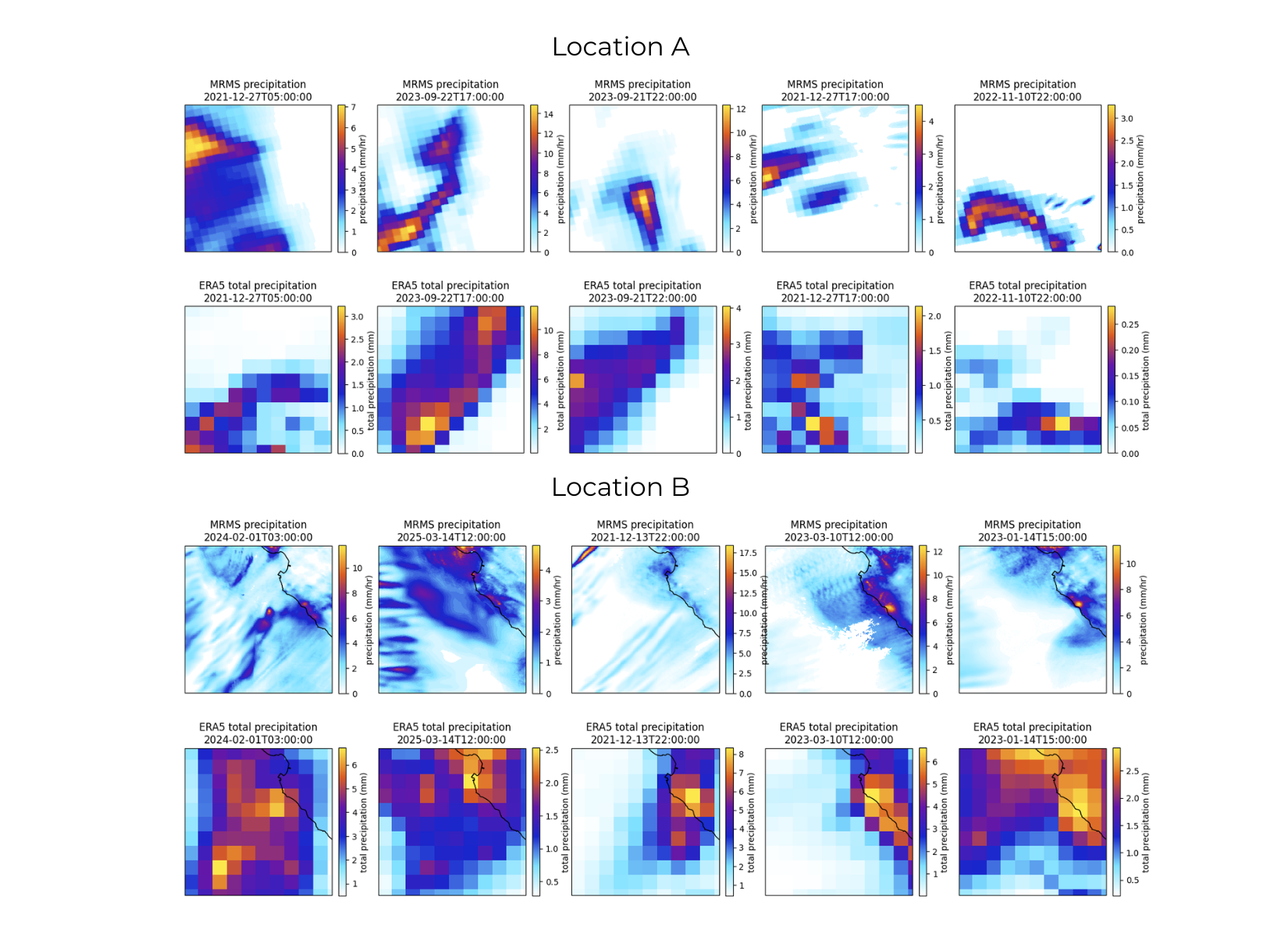}
  \caption{Comparison of MRMS and ERA5 precipitation samples. The high-resolution MRMS crops \textbf{(top)} are compared with the low-resolution ERA5 crops \textbf{(bottom)} for two different locations. In \textbf{Location A}, the MRMS data appears to be upsampled from a coarser grid. In contrast, \textbf{Location B} shows the typical fine-scale details expected from MRMS.}
\end{figure}

\paragraph{Masking MRMS for radar coverage} Across the CONUS domain, the MRMS dataset shows regional heterogeneities in effective resolution (see Figure \ref{fig:mrms-mr}). Further exploration revealed that some peripheral regions lack radar and rain gauge coverage. In these areas, the grid is instead filled with upsampled coarse satellite and NWP-derived data. To retain only genuine km-scale observations, we mask these regions following the spatial coverage of the variable \texttt{PrecipFlag}, which indicates the type of precipitation (\textit{e.g.}, convective, stratiform, hail) from radar observations.

\begin{figure}
  \centering
  \label{fig:euradclim-max}
  \includegraphics[width=\textwidth]{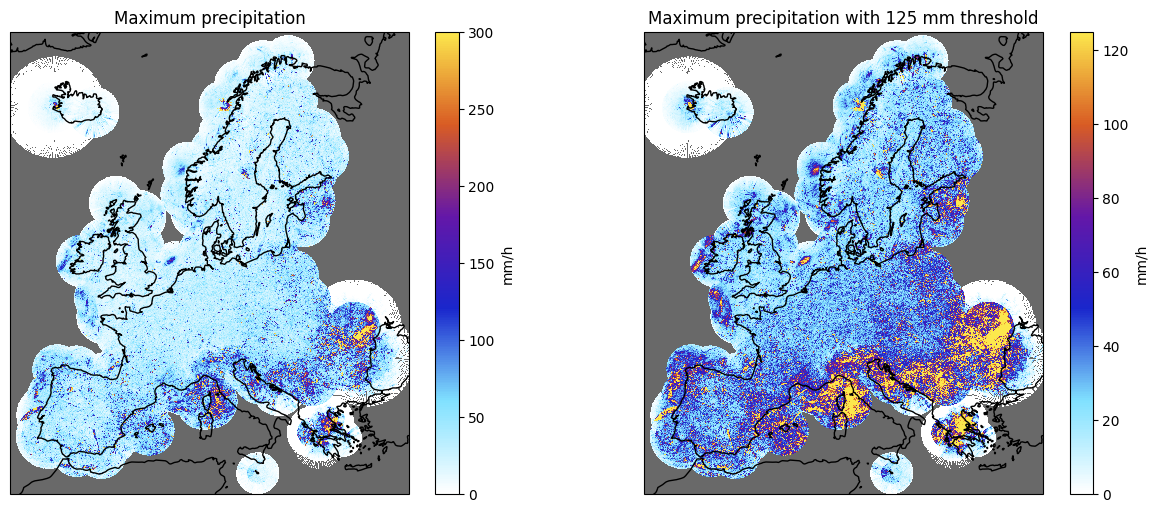}
  \caption{\textbf{(left)} Raw maximum hourly precipitation per grid-cell for EURADCLIM (2013 to 2023), and \textbf{(right)} same using a 125 \texttt{mm} threshold for easier identification of problematic grid-cells.}
\end{figure}

\paragraph{Incoherent extreme precipitation in EURADCLIM} Authors of the EURADCLIM dataset report that some regions might exhibit physically incoherent values, resulting from radar artifacts and sparse rain-gauge networks. In particular, the raw data contains hourly precipitation intensities reaching up to 300 \texttt{mm} (Figure \ref{fig:euradclim-max}, \textbf{left}), which is inconsistent with historical European extreme events records, which rarely exceed 150 \texttt{mm}\,/\,\texttt{h}. Setting a conservative threshold of 125 \texttt{mm}, we identified multiple clusters of affected grid-cells, especially over Norway, Estonia, Russia, Greece, Romania, Moldova, and some other coastal regions (Figure \ref{fig:euradclim-max} \textbf{right}). Because only $\sim 5\%$ of the total timesteps contained at least one grid cell exceeding this threshold, we discarded these timesteps entirely from our dataset.

\paragraph{Individual artifacts in EURADCLIM and EA} We removed the first 7 hours of 29 April 2019 from EURADCLIM due to a prominent spatial artifact over Spain. Similarly, for EA, we identified linear grid-line artifacts with unphysically large values occurring between 5 May and 7 May 2019, discarding the 26 affected timesteps from the final dataset.

\paragraph{Spatio-temporal alignment of ERA5 and high-resolution datasets} After processing the high-resolution precipitation source datasets, we employ an automated pipeline to retrieve and align the corresponding ERA5 reanalysis data. To maintain spatial alignment, we compute the geographic bounding box of the projected high-resolution domain, incorporating a coordinate buffer to ensure complete spatial coverage upon re-projection to a common equal-area projection. Regarding temporal alignment, we use hourly accumulated precipitation for low- and high-resolution precipitation data, and discard any timestamp that isn't include both in ERA5 and the high-resolution source datasets.

\paragraph{Storage and format} Due to the large spatial extent and km-scale resolution of our domains, the raw data requires roughly a dozen terabytes of memory before re-projection and resampling. To handle this volume, we rely heavily on parallelization and lazy computation. We process the data incrementally and write it to disk in chunks along the temporal dimension using the \textit{Zarr} format, which provides efficient compression. While the three source observational datasets would require 7.32 terabytes after preprocessing, each final \textit{Zarr} store needs only 150~GB.

\section{Stochastic spatio-temporal subsampling of precipitation crops}
\label{sec:appendix-subsampling}

Precipitation observations are largely dominated by dry conditions. To illustrate the effect of our subsampling methodology, Table~\ref{tab:statistics} compares the distributional statistics of $150\text{k}$ training crops extracted via uniform random sampling against rain-rate saturation importance sampling strategy proposed by \citet{dgmr}, across MRMS, EURADCLIM, and EA. Under uniform random sampling, dry grid cells ($\le 0$~mm) account for $82.8\%$ to $96.7\%$ of observations, while fewer than $1\%$ of spatial crops in MRMS and EA contain more than $50\%$ wet grid cells. In contrast, importance sampling significantly enhances the representation of heavy rainfall regimes: it increases the average crop intensity by up to $7\times$, elevates the average maximum intensity up to $19.84$~mm, and boosts the proportion of crops with more than $50\%$ wet grid cells by more than an order of magnitude.

\begin{table}[h]
    \caption{Statistics of \texttt{RainAtlas}'s training datasets with $150$k samples, generated with random or importance subsampling. The first section contains grid cell occurrences of multiple precipitation intensities, while the second section gives information on crops: average precipitation, average maximum intensity, and proportion of crops with more than $10\%, 25\%$ and $50\%$ non-zero precipitation grid cells.}
    \label{tab:statistics}
    \centering
    \begin{tabular}{rcccccc} 
        \toprule
        & \multicolumn{2}{c}{MRMS} & \multicolumn{2}{c}{EURADCLIM} & \multicolumn{2}{c}{EA} \\ \cmidrule(r){2-3} \cmidrule(r){4-5} \cmidrule(r){6-7} 
        & Random & Importance & Random & Importance & Random & Importance \\ \midrule
        $\leq 0$ mm (\%) & 95.80 & 72.76 & 82.81 & 55.77 & 96.68 & 80.33 \\
        $0 - 1$ mm (\%) & 2.38 & 12.63 & 15.34 & 37.18 & 1.01 & 5.05\\
        $1 - 4$ mm (\%) & 1.47 & 11.72 & 1.65 & 7.13 & 1.77 & 10.58 \\
        $4 - 5$ mm (\%) & 0.10 & 0.91 & 0.08 & 0.36 & 0.15 & 1.03 \\
        $5 - 8$ mm (\%) & 0.13 & 1.14 & 0.08 & 0.39 & 0.19 & 1.37 \\
        $8 - 10$ mm (\%) & 0.04 & 0.29 & 0.02 & 0.08 & 0.06 & 0.44 \\
        $> 10$ mm (\%) & 0.08 & 0.55 & 0.02 & 0.09 & 0.15 & 1.19 \\ \midrule
        average (mm) & 0.07 & 0.55 & 0.07 & 0.29 & 0.09 & 0.66 \\
        maximum (mm) & 6.53 & 19.84 & 5.83 & 13.10 & 4.61 & 16.60 \\
        nz $> 10$ (\%) & 13.30 & 78.10 & 48.47 & 95.25 & 10.33 & 57.64 \\
        nz $> 25$ (\%) & 4.58 & 45.99 & 27.27 & 78.55 & 3.66 & 31.39 \\
        nz $> 50$ (\%) & 0.88 & 14.47 & 8.89 & 41.10 & 0.75 & 9.55 \\
        \bottomrule
    \end{tabular}
\end{table}

\section{Architectures and training}
\label{sec:appendix-architectures}

\paragraph{Transformation and normalization} With the exception of the static land-sea mask, all variables undergo a fourth-root transformation ($y = x^{0.25}$) to mitigate skewness, and are then standardized independently for each dataset using $z$-score normalization.

\subsection{Training details}

All models are trained using 187,000 steps with a batch-size of 32 samples and the AdamW \citet{adamw} optimization algorithm. Learning rates follow a linear warm-up from $1~\%$ to $100~\%$ of the base learning rate over the first $2~\%$ of steps, followed by cosine annealing to $\eta_{\min}=10^{-6}$ over the remaining $98~\%$. Mixed precision is used for all models. Gradients are clipped so that their $L_2$ norm doesn't exceed 1.0. Following their original implementation, SerpentFlow and Consistency are trained using Exponential Moving Average (EMA). Each configuration is trained with three random seeds ($42$, $84$, $126$). Due to time constraints, a few configurations are evaluated over only one or two random seeds. Results will be updated during the rebuttal period. Individual hyperparameters are defined in Table \ref{tab:appendix-hyperparams}.

\begin{figure}
    \centering
    \includegraphics[width=1.0\linewidth]{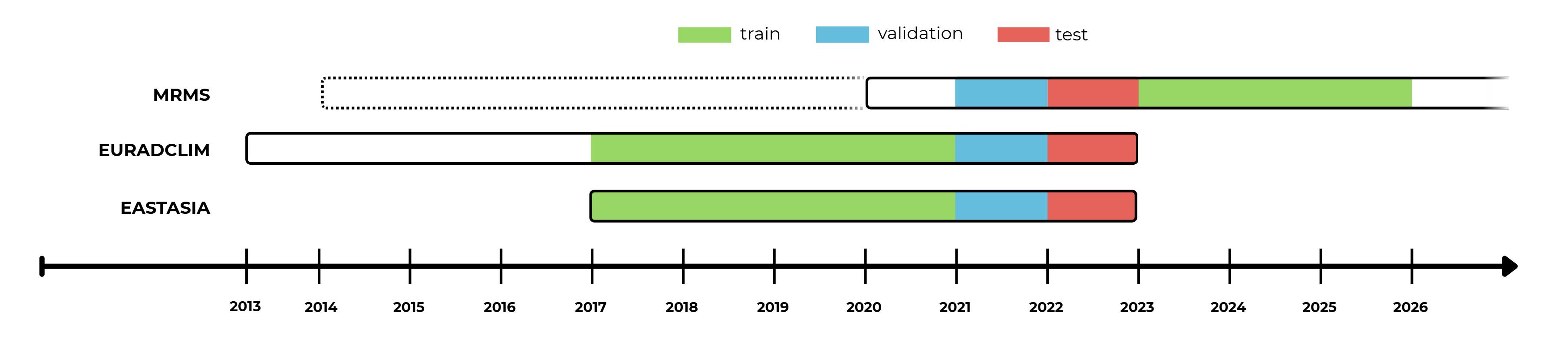}
    \caption{\textbf{Train, validation, and test temporal splits.} Because MRMS only partially overlap with EURADLCIM and EASTASIA, we reserve 2021 and 2022 for subsampling validation and test datasets. Since atmospheric predictability is limited to approximately 14 days, training on data from later periods does not induce data leakage.}
    \label{fig:train-val-test-splits-years}
\end{figure}

\subsection{Deterministic UNet}

The UNet backbone shared by all models is the EDM UNet of \citet{karras2022edm} adapted to climate downscaling by \citep{watt2024generative}. We use $64$ base channels, a per-resolution multiplier of $[1, 2, 3, 4]$ over four spatial stages ($256 \to 128 \to 64 \to 32$), two encoder blocks plus three decoder blocks per stage, and decoder-only self-attention at the three deepest stages ($32{\times}32$, $16{\times}16$, $8{\times}8$). The deterministic UNet baseline is trained on the standardized residual so that $\hat y = \mathrm{UNet}(x) + \bar x$, where $\bar x$ is the bilinearly upsampled ERA5 total-precipitation.

\subsection{Diffusion}

The diffusion baseline trains the EDM denoiser of \citet{karras2022edm} directly on the precipitation target conditioned on the bilinearly upsampled ERA5 predictors and the static covariates. The denoiser shares the UNet backbone wrapped by the EDM preconditioner. We use the EDM training noise schedule with $P_{\mathrm{mean}}=-1.2$, $P_{\mathrm{std}}=1.2$, and $\sigma_{\mathrm{data}}=0.5$, as well as the EDM denoising loss \citep{karras2022edm}. 

At inference we run the EDM Heun sampler \citet{karras2022edm} with $100$ second-order steps for $8$ ensemble members, $0.002$, $\sigma_{\max}=80$, $\rho=7$, and stochastic churn parameters $S_{\mathrm{churn}}=40$, $S_{\min}=0$, $S_{\max}=\infty$, $S_{\mathrm{noise}}=1$. 

\subsection{\textit{Corrective} Diffusion}

The \textit{corrective} diffusion baseline is adapted from CorrDiff \citet{mardani2025-corrdiff}, decomposing the downscaling task into a deterministic mean prediction followed by residual generative modeling. First, a deterministic UNet predicts the conditional mean $\mu(x) = \mathbb{E}[y \mid x]$, where $y$ denotes the target km-scale precipitation field and $x$ denotes the ERA5 and static conditioning variables. Second, a conditional diffusion model is trained to model the residual distribution $r = y - \mu(x)$, capturing high-frequency spatial details and unresolved sub-grid variability.

The UNet conditional mean regression stage reuses the identical setting as the UNet deterministic baseline, while the generative stage shares the diffusion baseline setting.

\subsection{SerpentFlow}

SerpentFlow is a two-stage downscaling generative model designed for unpaired domain alignment \citep{keisler2026serpentflow}. Given low- and high-resolution domains, respectively $\mathcal{D}_A$ and $\mathcal{D}_B$, SerpentFlow assumes that there exists a shared latent space $\mathcal{Z}$ and a bijective mapping $\mu$ such that: $\mu(\mathcal{D}_A) = \mathcal{B}_A \subset \mathcal{Z}$ and $\mu(\mathcal{D}_B) = \mathcal{B}_B \subset \mathcal{Z}$. Further, both the target and input latent domains can be decomposed into shared and specific components: $\mathcal{B}_A = \mathcal{B}^S \oplus \mathcal{B}^D_A$ and $\mathcal{B}_B = \mathcal{B}^S \oplus \mathcal{B}^D_B$. 

Instead of training on paired low- and high-resolution samples, SerpentFlow constructs pseudo-pairs from the target samples by replacing their latent specific components with white noise. Given a target sample $y^i$, $\epsilon \sim \mathcal{N}(0,1)$, and its transformation into the shared latent space $\mu(\epsilon) = z^S + z^D_{\epsilon} \in \mathcal{Z}^S \oplus \mathcal{Z}^D_{\epsilon} \subset \mathcal{Z}$, we construct the pseudo-view $\tilde{y}^i$ such that:
\begin{equation}
    (\tilde{y}^i, y^i) \in (\mu^{-1}(\mathcal{B}^S \oplus \mathcal{B}^D_{\epsilon}) \times \mathcal{D}_B), \quad \text{with } \tilde{y}^i = \mu^{-1}(z^{i, S}_B + z^D_{\epsilon}),
\end{equation}
where $z^{i, S}_B = \mu(y^i)^S$ represents the shared latent component of the target sample.

An informed choice for $\mathcal{Z}$ is the Fourier domain, where low-frequency signals correspond to large-scale synoptic processes shared by low- and high-resolution precipitation datasets. Using a cut-off frequency $\omega_c$, SerpentFlow constructs the shared and specific elements of the latent space $\mathcal{Z}$ following: 
\begin{equation}
    \mathcal{B}^S = \{z(\xi) : \|\xi\| < \omega_c\}, \quad \mathcal{B}^D = \{z(\xi) : \|\xi\| \geq \omega_c\},
\end{equation}
where $\xi$ represents the frequency components in the Fourier domain.

SerpentFlow proposes an automatic method for selecting the cut-off frequency. Given a candidate $\omega_c$, a low-pass filter is applied to each sample from the input and target domains:
\begin{equation}
    x^S = \mathcal{F}^{-1}[\mathds{1}_{\{\|\xi\| < \omega_c\}} \cdot \mathcal{F}(x)]
\end{equation}
where $\mathcal{F}$ is the Fourier transform. Starting from a high candidate cut-off frequency, SerpentFlow trains a classifier $D_{\psi}$ to predict whether $x^S$ originates from $\mathcal{D}_A$ or $\mathcal{D}_B$ using the standard binary cross-entropy loss. $\omega_c$ is decreased iteratively until the validation accuracy of $D_{\psi}$ reaches an indiscriminable level: 
\begin{equation}
    Acc(D_{\psi}, \omega_c^*) \approx 0.5,
\end{equation}
which indicates that there is no more specific information contained in the samples given the optimal cut-off frequency $\omega_c^*$.

We use the same 3-layer convolutional neural network as \citet{keisler2026serpentflow} for the classifier $D_{\psi}$. Once $\omega_c^*$ is reached, we train a flow-matching model $f_{\theta}$ for reconstructing $y^i$ from $\tilde{y}^i_{\omega_c^*}$ following the same setting as the SerpentFlow authors. At inference, we sample the low-resolution precipitation field $x^i \in \mathcal{D}_A$ and predict the corresponding high-resolution target $\hat{y}^i = f_{\theta}(\tilde{x}^i_{\omega_c^*})$ using the flow-matching model.

\subsection{Consistency}

The consistency baseline reproduces the continuous-time framework of \citet{song2023consistency}, which is designed to overcome the slow, iterative sampling process of traditional diffusion models. We utilize the identical EDM UNet backbone and conditioning strategy on the bilinearly upsampled ERA5 predictors and static covariates as our diffusion baseline.

Building upon this continuous-time formulation, consistency models learn a mapping $f_{\theta}(y_t, t, x)$ that points from any intermediate noisy state $y_t$ directly back to its clean origin $y$. This requires the model to satisfy the consistency property, meaning that evaluations at any two arbitrary time steps $t$ and $t'$ along the same diffusion trajectory must yield the same prediction:
\begin{equation}
    f_{\theta}(y_t, t, x) = f_{\theta}(y_{t'}, t', x), \quad \forall t, t' \in [\epsilon, T].
\end{equation}

A critical requirement is the boundary condition $f_{\theta}(y_{\epsilon}, \epsilon, x) = y_{\epsilon}$. To strictly enforce this, the model prediction is parameterized using modified time-dependent skip connections:
\begin{equation}
    f_{\theta}(y_t, t, x) = c_{\text{skip}}(t) y_t + c_{\text{out}}(t) F_{\theta}(y_t, t, x),
\end{equation}
where $F_{\theta}$ is the raw neural network output. Unlike the standard EDM preconditioner, the consistency weighting functions are explicitly constrained such that $c_{\text{skip}}(\epsilon) = 1$ and $c_{\text{out}}(\epsilon) = 0$, ensuring the model perfectly returns the input at the minimum noise level $\epsilon$.

To enforce self-consistency across the trajectory, the model minimizes the discrepancy between predictions made at adjacent time steps $(t_{n+1}, t_n)$ derived from the discretized EDM noise schedule:
\begin{equation}
    \mathcal{L}(\theta, \theta^-) = \mathbb{E}_{y, x, n, z} \left[\lambda(\sigma_i) d\left( f_{\theta}(y_{t_{n+1}}, t_{n+1}, x), f_{\theta^-}(y_{t_n}, t_n, x) \right) \right],
\end{equation}
where $\theta^-$ represents the exponential moving average (EMA) of the online network parameters $\theta$, $\lambda(\sigma_i)$ weights the relative contribution of different noise levels, $d(\cdot, \cdot)$ is a distance metric, and $z \sim \mathcal{N}(0, I)$ is the noise applied to sample $y_{t_{n+1}}$ and $y_{t_n}$. This EMA target network stabilizes training and ensures a continuous learning trajectory relative to the online parameters. Following recommendations from \citet{song2024improved}, we set the EMA decay parameter to 0, formally fixing $\theta = \theta^-$. This provably eliminates the inherent bias and approximation error introduced by a lagging target network. 

We adopt several additional techniques from \citet{song2024improved} to improve performance. First, we replace the $L_2$ distance in the consistency loss $\mathcal{L}$ with the Pseudo-Huber metric: $d(x,y) = \sqrt{\|x - y\|_2^2 + c^2} - c$. Following the authors, we fix $c = 0.00054\sqrt{d}$ for samples of dimension $d$ (we note this hyperparameter was originally optimized for natural images and may be suboptimal for precipitation fields). Second, we replace the uniform weighting function $\forall \, i, \lambda(\sigma_i) = 1$ with $\lambda(\sigma_i) = \frac{1}{\sigma_{i+1} - \sigma_i}$, and we implement the proposed discretized lognormal noise schedule. Finally, we set the minimum number of discretization bins to 10, and for reasons of training stability, we lower the number of maximal bins to 50.

At inference, starting from pure Gaussian noise $y_T \sim \mathcal{N}(0, T^2 I)$ and the given conditioning variables $x$, the model predicts the corresponding high-resolution precipitation field $\hat{y}$ in a single forward pass:
\begin{equation}
    \hat{y} = f_{\theta}(y_T, T, x).
\end{equation}

We initially sought to reproduce the unpaired approach proposed by \citep{hess-consistency}. This method relies on finding the wavenumber $k^*$ where the Power Spectral Densities (PSDs) of the low- and high-resolution datasets intersect to determine the corresponding noise timestep $t^*$. Inference is then performed from the perturbed state $\tilde{x} = x + \epsilon_{t^*}$, with $\epsilon_{t^*} \sim \mathcal{N}(0, \sigma^2(t^*)\mathbf{I})$ and $\sigma(t) = N^2\text{PSD}(k)$ for grid size $N$. However, because the PSDs of our low- and high-resolution datasets do not intersect at any wavenumber, applying this specific unpaired strategy proved impossible.

\begin{table}[!htbp]
  \caption{Training and architecture hyperparemeters for every model. Additional hyperparameters are specified under each model specification when they are not shared with others. We note that most of these hyperparameters are certainly sub-optimals as we didn't perform any extensive hyperparameters seach. Most values are taken from previous work \citep{watt2024generative, karras2022edm, song2024improved, keisler2026serpentflow}.}
  \label{tab:appendix-hyperparams}
  \centering
  \footnotesize
  \setlength{\tabcolsep}{4pt}
  \resizebox{\textwidth}{!}{
  \begin{tabular}{lccccc}
    \toprule
    & UNet & Diffusion & \textit{Corrective} Diffusion & SerpentFlow & Consistency \\
    \midrule
    Loss & $L_2$ & EDM denoising & EDM denoising & $L_2$ & Pseudo-Huber \\
    Optimizer & AdamW & AdamW & AdamW & AdamW & AdamW\\
    Learning rate & $2{\times}10^{-4}$ & $10^{-4}$ & $10^{-4}$ & $10^{-4}$ & $10^{-4}$ \\
    Weight decay & $10^{-4}$ & $10^{-2}$ & $10^{-2}$ & $10^{-2}$ & $10^{-2}$ \\
    \midrule
    Residual & \checkmark & --- & \checkmark (UNet) & --- & --- \\
    Dropout & 0.1 & 0.1 & 0.1 & 0.1 & 0.1 \\
    EDM $(P_{\mathrm{mean}},P_{\mathrm{std}},\sigma_{\mathrm{data}})$ & --- & $(-1.2,\,1.2,\,0.5)$ & $(-1.2,\,1.2,\,0.5)$ & --- & $(-1.1,\,2.0,\,0.5)$ \\
    EDM $(\sigma_{\min},\sigma_{\max},\rho)$ & --- & $(0.002,\,80,\,7)$ & $(0.002,\,80,\,7)$ & --- & $(0.002,\,80,\,7)$ \\
    EMA decay & --- & --- & --- & 0.999 & 0 \\
    ODE solver & --- & Heun & Heun & dopri5 & --- \\
    Sampling steps or $(r_{tol} / a_{tol})$ & --- & $100$ & $100$ & $(10^{-3}/10^{-4})$ & 1 \\
    \bottomrule
  \end{tabular}
  }
\end{table}

\section{Metrics}
\label{sec:appendix-metrics}

Let $y \in \mathbb{R}^{H \times W}$ denote the ground-truth high-resolution precipitation field ($H = W = 256$), and let $\hat{y} \in \mathbb{R}^{H \times W}$ denote the prediction from a deterministic baseline. For generative baselines, models generate an ensemble of $M$ realizations, denoted by $\{\hat{y}^{(m)}\}_{m=1}^M$ ($M = 8$ across all experiments), with empirical ensemble mean $\bar{y} = \frac{1}{M} \sum_{m=1}^M \hat{y}^{(m)}$.

\subsection{Point-wise and probabilistic errors}

To determine whether predictions systematically under- or over-estimate precipitation, we compute the mean BIAS. For deterministic baselines, point-wise error is measured via the Mean Absolute Error (MAE):
\begin{equation}
    \text{MAE}(y, \hat{y}) = \frac{1}{|\Omega|} \sum_{i \in \Omega} |y_i - \hat{y}_i| , \qquad
    \text{BIAS}(y, \hat{y}) = \frac{1}{|\Omega|} \sum_{i \in \Omega} (\hat{y}_i - y_i) ,
\end{equation}
where $\Omega$ denotes the set of grid cells with non-missing values within the spatial domain. For generative models, the BIAS is computed by substituting $\hat{y}$ for the ensemble mean $\bar{y}$. 

For generative models, point-wise errors are evaluated using the MAE's probabilistic generalization: the Continuous Ranked Probability Score (CRPS) \citep{hans2000-crps}. Since computing the CRPS in its integral form is intractable with a limited number of ensemble members, it is evaluated pointwise at each grid cell $i$ via its energy form representation, given by:
\begin{equation}
    \text{CRPS}(y, \{\hat{y}^{(m)}\}_{m=1}^M) = \frac{1}{|\Omega|} \sum_{i \in \Omega} \left( \frac{1}{M}\sum_{m=1}^M |\hat{y}_i^{(m)} - y_i| - \frac{1}{2M(M-1)}\sum_{m=1}^M \sum_{n=1}^M |\hat{y}_i^{(m)} - \hat{y}_i^{(n)}| \right) .
\end{equation}
We use this fair estimator, which is unbiased for the CRPS of the distribution the ensemble is drawn from \citep{ferro2014-fair}. For a deterministic prediction ($M = 1$), the second term of the CRPS vanishes and the score reduces to the MAE. Therefore, we compare the MAE of deterministic models with the CRPS of generative models. To evaluate sensitivity to the double-penalty problem, BIAS, MAE and CRPS are also computed after spatially pooling fields with average pooling filters of stride and kernel sizes $k \in \{4, 8, 12\}$.

\subsection{Scale-dependent skill}

The fractions skill score (FSS) \citet{roberts2008scale} compares the fraction of grid cells above a threshold $q$ within a square neighbourhood of size $n$ in the prediction, $f_i$, and in the observation, $o_i$:
\begin{equation}
    \text{FSS}_{q,n} = 1 - \frac{\sum_{i \in \Omega} (f_i - o_i)^2}{\sum_{i \in \Omega} f_i^2 + \sum_{i \in \Omega} o_i^2} .
\end{equation}
We use $q \in \{1, 5, 10, 20\}$ mm\,h$^{-1}$ and $n \in \{2, 4, 8, 16, 32\}$ grid cells. For generative models, $f_i$ is computed from the ensemble mean. The numerator and denominator are summed over all test samples before taking the ratio, so samples with no exceedance of $q$ in either field do not contribute. FSS ranges from $0$ to $1$, and higher is better.

\subsection{Spatial structure}

We evaluate the variogram score of order $p$ to assess whether generated high-resolution fields reproduce observed spatial variability, local gradients, and sub-grid structure across scales \citep{variogram-score}. Unlike point-wise metrics, the variogram score is computed through pair-wise differences, making it a more robust scoring rule against the double-penalty problem:
\begin{equation}
    \text{VS}_p(y, \{\hat{y}^{(m)}\}_{m=1}^M) = \frac{1}{\sum_{i \in \Omega} \sum_{j \in \Omega_i} w_{i,j}} \sum_{i \in \Omega} \sum_{j \in \Omega_i} w_{i,j} \left( |y_i - y_j|^p - \frac{1}{M} \sum_{m=1}^M |\hat{y}_i^{(m)} - \hat{y}_j^{(m)}|^p \right)^2 ,
\end{equation}
where $\Omega_i = \{j \in \Omega : 0 < \|r_i - r_j\|_2 \le r_{\max}\}$ defines the neighbourhood of grid cell $i$ up to a maximum separation radius $r_{\max}$. Following standard meteorological practice, we set $p = 0.5$ and use an inverse Euclidean distance weighting scheme, $w_{i,j} = \frac{1}{\|r_i - r_j\|_2}$, evaluated over separation distances up to $r_{\max} = 10\text{ km}$ ($5$ grid units on the $2\text{ km}$ target grid).

\subsection{Spectral and intensity fidelity}

Since high-intensity precipitation events are not well represented in the input \textit{total precipitation} variable from the coarse ERA5 reanalysis, we assess how baselines reconstruct the ground-truth intensity distribution by computing intensity histograms, and the resulting Logarithmic Histogram Distance (LHD). Additionally, km-scale observed precipitation has much finer spatial details than ERA5, often combined with intense precipitation, which results in higher spatial frequencies. To evaluate whether models reproduce these frequencies without introducing spectral artifacts, we compute the Radially Averaged Power Spectral Density (RAPSD).

\paragraph{Intensity histograms and Logarithmic Histogram Distance (LHD).}

Intensity histograms are computed over $B$ uniformly spaced bins between 0 and 300 \texttt{mm}, with a length of 1 \texttt{mm} per bin (\textit{i.e.,} $B = 300$). Because precipitation distributions are heavy-tailed and dominated by dry grid cells, standard linear probability distances are heavily skewed toward low intensities and obscure errors in heavy rainfall regimes. To benchmark distributional fidelity across both light rain and rare extreme events, we compute the Logarithmic Histogram Distance (LHD). We define $\mathcal{B}_V = \{b : c_b + \hat{c}_b) > 10\}$ as the set of valid bins for computing the LHD, where $c_b$ and $\hat{c}_b$ are the respective counts in bin $b$ for the ground-truth and predictions. Let $p_b(y)$ and $p_b(\hat{y})$ denote the empirical probability mass of observation $y$ and prediction $\hat{y}$ falling into valid bin $b \in \mathcal{B}_V$, normalized over grid cells with non-missing values $\Omega$. The LHD is defined as:
\begin{equation}
    \text{LHD}(y, \hat{y}) = \sqrt{ \frac{1}{|\mathcal{B}_V |} \sum_{b \in \mathcal{B}_V} (\log_{10}p_b(\hat{y} + \epsilon) - \log_{10}(p_b(y) + \epsilon))^2 } ,
\end{equation}
where $\epsilon = 10^{-16}$ is a small regularization constant to prevent undefined logarithms for empty bins. For generative models, $p_b(\hat{y})$ is computed by summing counts across all $M$ ensemble members prior to calculating the bin frequencies.

\paragraph{Radially Averaged Power Spectral Density (RAPSD).}
We compute the RAPSD \citet{ruz2011-rapsd} to evaluate whether models reproduce the spatial frequencies observed in high-resolution observations, thereby producing realistic km-scale precipitation fields. To avoid introducing high-frequency spectral artifacts from zero-padding or imputation masks, spatial crops containing any missing values are excluded entirely from the spectral evaluation. Given a spatial field $x \in \mathbb{R}^{H \times W}$, we compute its two-dimensional discrete Fourier transform:
\begin{equation}
    \hat{X}(k_u, k_v) = \sum_{u=0}^{H-1} \sum_{v=0}^{W-1} x(u, v) \exp\left( -2\pi i \left( \frac{k_u u}{H} + \frac{k_v v}{W} \right) \right) ,
\end{equation}
yielding the two-dimensional power spectrum $P(k_u, k_v) = |\hat{X}(k_u, k_v)|^2$. The radially averaged power spectrum $E(k)$ is obtained by integrating $P(k_u, k_v)$ over concentric annuli in wavenumber space:
\begin{equation}
    E(k) = \frac{1}{|\mathcal{A}_k|} \sum_{(k_u, k_v) \in \mathcal{A}_k} P(k_u, k_v) , \quad \mathcal{A}_k = \left\{ (k_u, k_v) : k - \frac{\Delta k}{2} \le \sqrt{k_u^2 + k_v^2} < k + \frac{\Delta k}{2} \right\} ,
\end{equation}
where $k$ represents the isotropic radial wavenumber and $\Delta k = 1$. The radial wavenumbers correspond to physical spatial wavelengths $\lambda = \frac{L}{k}$, ranging from the domain extent ($L = 512\text{ km}$ at $k = 1$) down to the Nyquist limit ($\lambda = 4\text{ km}$ at $k = 128$). For generative baselines, $E(k)$ is computed independently for each ensemble member and averaged across members and evaluation crops.

\subsection{Calibration and uncertainty}

Evaluating model calibration is critical to ensure that the true uncertainty of the physical process is accurately captured by the spread of the ensemble predictions. To quantify dispersion and compare generative baselines using a single scalar metric, we compute the Spread-Skill Ratio (SSR). However, because the SSR does not reveal the underlying structure of this dispersion, we additionally construct Talagrand rank histograms \citep{hamill2001-rankhistograms}. By evaluating where observations fall within the ensemble's uncertainty distribution, these histograms allow us to diagnose exactly how and where the model's confidence deviates from the true probabilities.

\paragraph{Spread-Skill Ratio (SSR).}
The SSR compares the internal model uncertainty against the prediction error. For a perfectly calibrated ensemble, the ensemble spread should equal the Root Mean Square Error (RMSE) of the ensemble mean. The SSR is defined as:
\begin{equation}
    \text{SSR} = \sqrt{\frac{M+1}{M}} \, \frac{ \sqrt{ \frac{1}{|\Omega|} \sum_{i \in \Omega} \frac{1}{M-1} \sum_{m=1}^M (\hat{y}_i^{(m)} - \bar{y}_i)^2 } } { \sqrt{ \frac{1}{|\Omega|} \sum_{i \in \Omega} (\bar{y}_i - y_i)^2 }}  ,
\end{equation}
where the numerator is the root mean ensemble variance, the factor $\sqrt{(M+1)/M}$ corrects for the finite ensemble size \citet{fortin2014-ssr}, and the denominator is the RMSE of the empirical ensemble mean $\bar{y}$. Variances and squared errors are pooled over all grid cells and test crops before taking the square roots. An $\text{SSR} \approx 1$ indicates a calibrated ensemble, whereas values strictly less than 1 indicate under-dispersion (over-confidence) and values greater than 1 indicate over-dispersion (under-confidence). 

% Note that to compute the ensemble spread, we use the mean standard deviation instead of the root mean variance. Due to Jensen's law ($\mathbb{E}[\sigma] \leq \mathbb{E}[\sigma^2]$), this results in an under-estimation of the numerator, and therefore a more conservative SSR (systematically more under-dispersed).

\paragraph{Rank Histograms.}

Talagrand rank histograms \citet{hamill2001-rankhistograms} evaluate whether ground-truth observations could originate from the predictive ensemble distribution. For each valid grid cell $i \in \Omega$, the observation $y_i$ is compared against the ensemble predictions $\{\hat{y}_i^{(m)}\}_{m=1}^M$. The rank $R_i \in \{1, \dots, M+1\}$ is determined by counting the number of ensemble members strictly less than the observation. However, since precipitation fields are heavily zero-inflated (see Table \ref{tab:statistics}), dry observations would often tie with several members but get classified under the first rank, producing artificial L-shape histograms. We set values below a threshold of 0.1 \texttt{mm} to zero in both predictions and observations. We therefore break ties at random \citet{rankhistogram-ties}:
\begin{equation}
     R_i = 1 + \sum_{m=1}^M \mathds{1}_{\{\hat{y}_i^{(m)} < y_i\}} + U_i , \quad U_i \sim \mathcal{U}\{0, \dots, T_i\} , \quad T_i = \sum_{m=1}^M \mathds{1}_{\{\hat{y}_i^{(m)} = y_i\}} ,
\end{equation}
where $\mathds{1}$ is the indicator function. By accumulating $R_i$ across all evaluation samples, we obtain a histogram of ranks. A perfectly calibrated model produces a uniform (flat) histogram. An over-confident model yields a $\cup$-shaped histogram (indicating the ensemble spread is too narrow to cover the true variance), while an over-dispersed model results in a $\cap$-shaped histogram. An L-shape histogram is the result of over-estimation of precipitation values across the entire ensemble spread. 

\section{Additional results}
\label{sec:appendix-results}

\subsection{In-domain performance}
\label{sec:appendix-indomain}

\subsubsection{Importance-subsampled test datasets}
\label{sec:appendix-indomain-importance}

\begin{figure}[!htbp]
    \centering
    \includegraphics[width=1.0\linewidth]{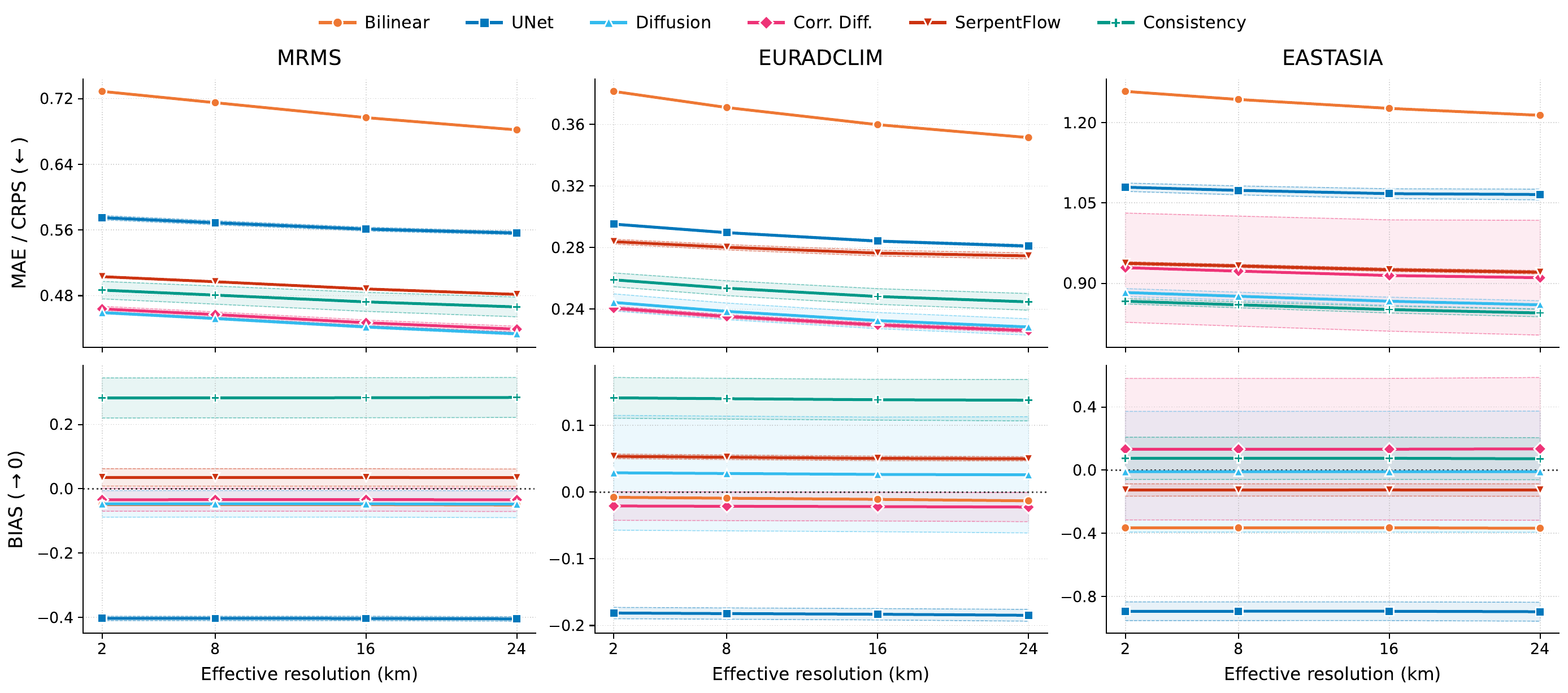}
    \caption{\textbf{Effect of spatial coarsening on pointwise error and bias (importance-subsampled test sets).} MAE / CRPS (\textbf{top}, $\downarrow$) and bias (\textbf{bottom}, $\rightarrow 0$) evaluated across four effective spatial resolutions (native $2$~km, $8$~km, $16$~km, and $24$~km) obtained by average pooling. Shaded bands denote $\pm$ standard deviation across three random seeds. Pointwise errors systematically decrease with spatial aggregation due to reduced sensitivity to small-scale displacement errors (double-penalty effect), while bias remains scale-invariant. Relative performance rankings remain unchanged across resolutions.}
    \label{fig:maecrpsbias-dsfs-importance}
\end{figure}

\begin{figure}[!htbp]
  \centering
  \includegraphics[width=1.0\textwidth]{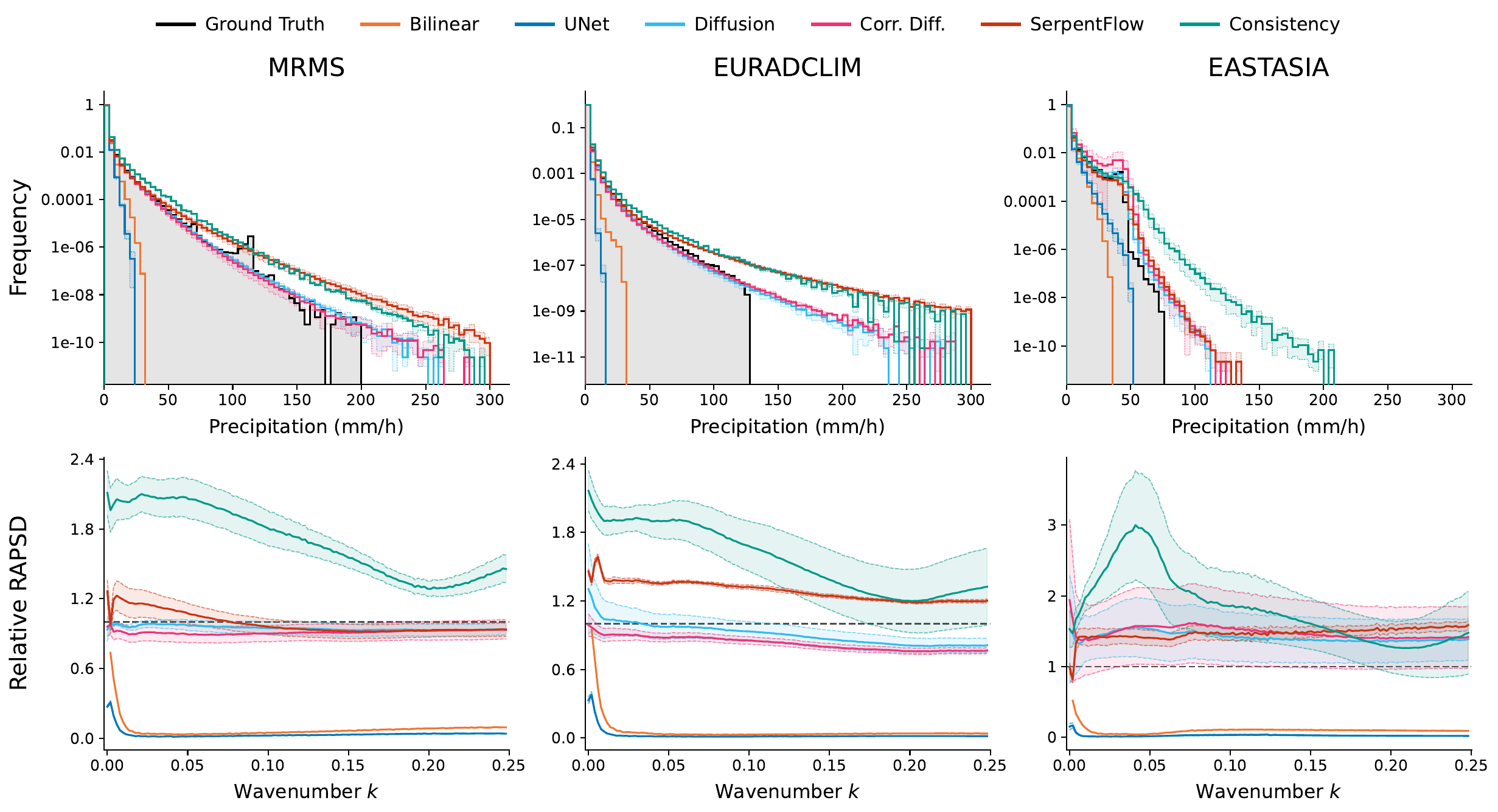}
  \caption{\textbf{In-domain intensity distribution and spectral fidelity (importance-subsampled test sets).} \textbf{(Top)} Precipitation intensity histograms comparing model predictions against ground truth up to $300$~mm\,h$^{-1}$ on a logarithmic scale. \textbf{(Bottom)} Radially Averaged Power Spectral Density (RAPSD) normalized by observed power spectrum across radial wavenumbers $k$. The horizontal dashed line at $y = 1.0$ indicates perfect spectral agreement with observations. Shaded bands represent $\pm$ standard deviation across evaluation crops and ensemble members.}
  \label{fig:indomain-histograms-rapsd-importance}
\end{figure}

\begin{figure}[!htbp]
    \centering
    \includegraphics[width=1.0\linewidth]{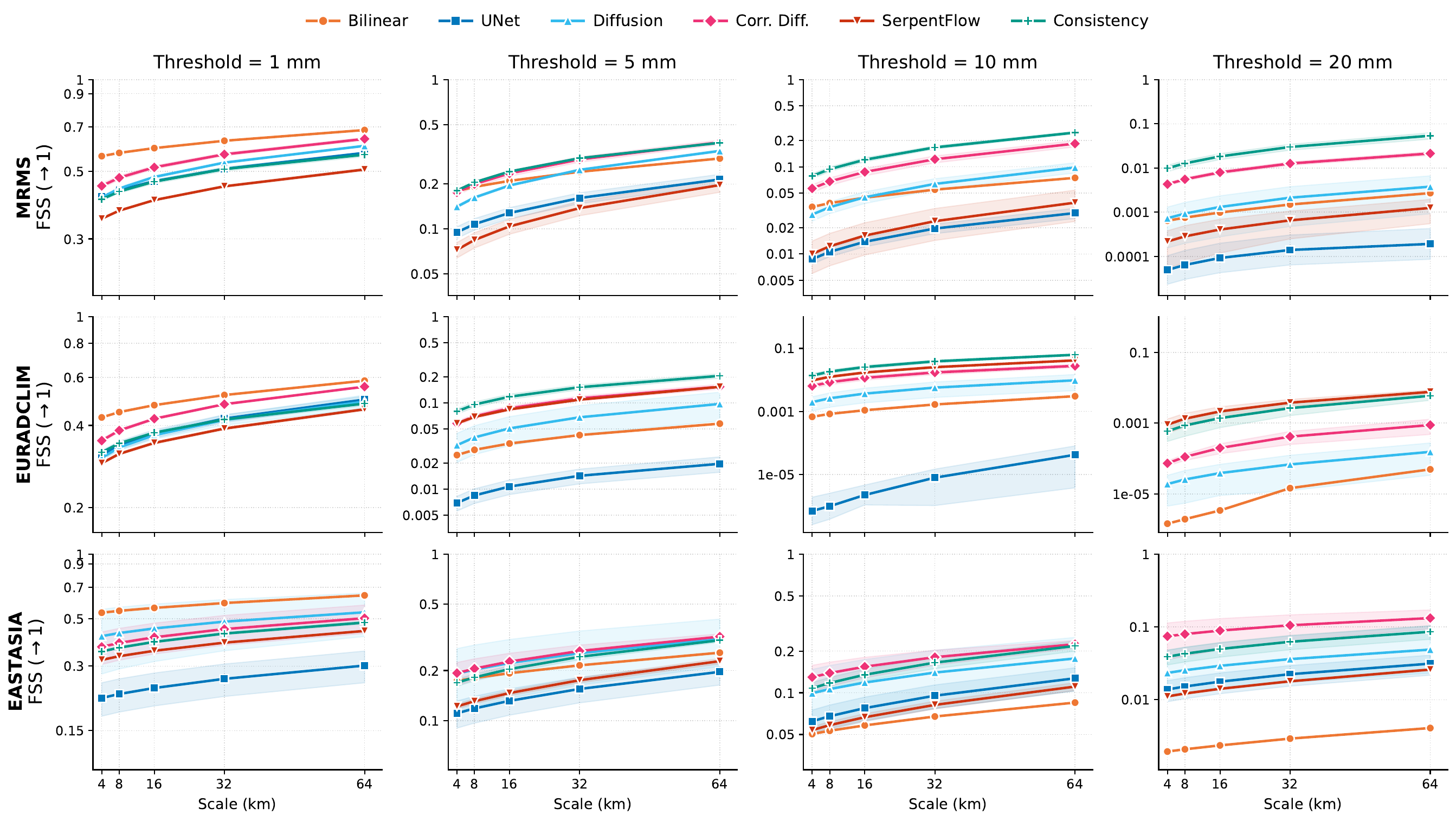}
    \caption{\textbf{Fractions Skill Score across precipitation thresholds (importance-subsampled test sets).} Fractions Skill Score (FSS, $\uparrow 1$) evaluated across neighborhood scales ($4, 8, 16, 32, 64$~km) for thresholds $q \in \{1, 5, 10, 20\}$~mm\,h$^{-1}$ (\textbf{left to right}). Shaded bands denote $\pm$ standard deviation across three random seeds. Generative baselines, especially diffusion and \textit{corrective} Diffusion, retain high skill for extreme precipitation ($q \ge 5$~mm\,h$^{-1}$), whereas deterministic baselines degrade sharply.}
    \label{fig:fss-importance}
\end{figure}

\begin{figure}[!htbp]
    \centering
    \includegraphics[width=1.0\linewidth]{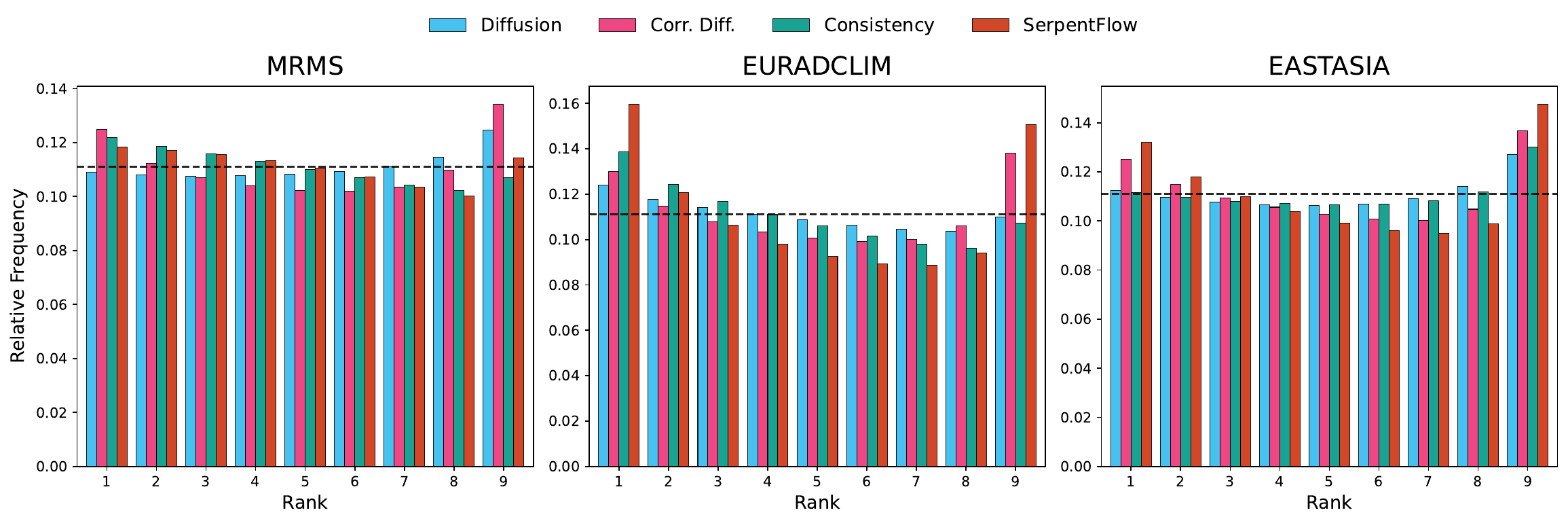}
    \caption{\textbf{Ensemble calibration via rank histograms (importance-subsampled test sets).} Talagrand rank histograms for generative models evaluated using an $8$-member ensemble ($9$ rank bins). The horizontal dashed line at $y \approx 0.111$ marks the ideal uniform distribution corresponding to perfect ensemble calibration. Random tie-breaking is applied to values below $0.1$~mm\,h$^{-1}$.}
    \label{fig:rank-histos-importance}
\end{figure}

We present comprehensive in-domain evaluation results on the importance-subsampled test datasets across all three continental domains (MRMS, EURADCLIM, and EASTASIA). Table~\ref{tab:indomain-results-importance} provides numerical values across all evaluated metrics. Figure~\ref{fig:maecrpsbias-dsfs-importance} examines the impact of spatial coarsening (from 2~km down to 24~km effective resolution) on pointwise errors and bias. Figure~\ref{fig:indomain-histograms-rapsd-importance} displays the intensity distribution and spectral fidelity via radially averaged power spectral density (RAPSD). Figure~\ref{fig:fss-importance} shows scale-dependent Fractions Skill Scores (FSS) across four precipitation thresholds. Finally, Figure~\ref{fig:rank-histos-importance} presents rank histograms evaluating ensemble calibration for the generative baselines.

\begin{table}[!htbp]
    \caption{\textbf{In-domain performance evaluation on importance-subsampled test datasets.} MAE, CRPS, and BIAS are computed on the original resolution without artificial coarsening. Probabilistic metrics (SSR and VARIOGRAM) are evaluated exclusively for generative models, omitting the deterministic Bilinear (ERA5) and UNet baselines. Arrows in the header indicate the optimal direction for each metric. The best and second-best results are \textbf{bolded} and \underline{underlined}, respectively. Subscripts denote standard deviations.}
    \label{tab:indomain-results-importance}
    \centering
    \resizebox{\textwidth}{!}{%
    \begin{tabular}{l l r@{\,\,}l r@{\,\,}l r@{\,\,}l r@{\,\,}l r@{\,\,}l} 
        \toprule
        & & \multicolumn{2}{c}{MAE / CRPS $(\downarrow)$} & \multicolumn{2}{c}{BIAS $(\rightarrow 0)$} & \multicolumn{2}{c}{LHD $(\downarrow)$} & \multicolumn{2}{c}{SSR $(\rightarrow 1)$} & \multicolumn{2}{c}{VARIOGRAM $(\downarrow)$} \\
        \midrule
        \multirow{6}{*}{MRMS} 
        & Bilinear                      & 0.7288 & & -0.0490 & & 19.9497 & & \multicolumn{2}{c}{-} & \multicolumn{2}{c}{-} \\            
        & UNet                          & 0.5750 & $_{\pm 0.0026}$ & -0.4033 & $_{\pm 0.0069}$ & 21.0799 & $_{\pm 0.1626}$ & \multicolumn{2}{c}{-} & \multicolumn{2}{c}{-} \\
        & Diffusion                     & \textbf{0.4593} & $_{\pm 0.0014}$ & -0.0472 & $_{\pm 0.0413}$ & \underline{4.4261} & $_{\pm 0.6763}$ & \underline{0.9437} & $_{\pm 0.0103}$ & \underline{0.1671} & $_{\pm 0.0004}$ \\
        & \textit{Corrective} Diffusion & \underline{0.4639} & $_{\pm 0.0031}$ & \textbf{-0.0341} & $_{\pm 0.0358}$ & \textbf{3.3621} & $_{\pm 1.0940}$ & 0.8373 & $_{\pm 0.0436}$ & \textbf{0.1662} & $_{\pm 0.0007}$ \\
        & SerpentFlow                   & 0.5031 & $_{\pm 0.0005}$ & \underline{0.0353} & $_{\pm 0.0271}$ & 8.5822 & $_{\pm 0.9519}$ & \textbf{1.0268} & $_{\pm 0.0616}$ & 0.1798 & $_{\pm 0.0012}$ \\
        & Consistency                   & 0.4867 & $_{\pm 0.0107}$ & 0.2825 & $_{\pm 0.0623}$ & 7.0539 & $_{\pm 0.2034}$ & 1.2294 & $_{\pm 0.0225}$ & 0.1825 & $_{\pm 0.0027}$ \\
        \midrule
        \multirow{6}{*}{EURADCLIM} 
        & Bilinear                      & 0.3815 & & \textbf{-0.0077} & & 18.5740 & & \multicolumn{2}{c}{-} & \multicolumn{2}{c}{-} \\
        & UNet                          & 0.2953 & $_{\pm 0.0007}$ & -0.1813 & $_{\pm 0.0085}$ & 21.1591 & $_{\pm 0.1674}$ & \multicolumn{2}{c}{-} & \multicolumn{2}{c}{-} \\
        & Diffusion                     & \underline{0.2444} & $_{\pm 0.0056}$ & 0.0290 & $_{\pm 0.0861}$ & \textbf{6.7102} & $_{\pm 0.8443}$ & \textbf{0.9124} & $_{\pm 0.0904}$ & \underline{0.0926} & $_{\pm 0.0020}$ \\
        & \textit{Corrective} Diffusion & \textbf{0.2407} & $_{\pm 0.0012}$ & \underline{-0.0208} & $_{\pm 0.0214}$ & \underline{7.4758} & $_{\pm 0.4153}$ & 0.7743 & $_{\pm 0.0192}$ & \textbf{0.0896} & $_{\pm 0.0002}$ \\
        & SerpentFlow                   & 0.2839 & $_{\pm 0.0016}$ & 0.0537 & $_{\pm 0.0034}$ & 12.4180 & $_{\pm 0.3756}$ & 0.8458 & $_{\pm 0.0139}$ & 0.1048 & $_{\pm 0.0011}$ \\
        & Consistency                   & 0.2591 & $_{\pm 0.0045}$ & 0.1414 & $_{\pm 0.0306}$ & 10.8817 & $_{\pm 0.2944}$ & \underline{1.1105} & $_{\pm 0.0268}$ & 0.0989 & $_{\pm 0.0025}$ \\
        \midrule
        \multirow{6}{*}{EA}   
        & Bilinear                      & 1.2581 & & -0.3655 & & 16.2220 & & \multicolumn{2}{c}{-} & \multicolumn{2}{c}{-} \\
        & UNet                          & 1.0794 & $_{\pm 0.0077}$ & -0.8937 & $_{\pm 0.0591}$ & 11.2942 & $_{\pm 0.8373}$ & \multicolumn{2}{c}{-} & \multicolumn{2}{c}{-} \\
        & Diffusion                     & \underline{0.8829} & $_{\pm 0.0074}$ & \textbf{-0.0099} & $_{\pm 0.3823}$ & \underline{6.5990} & $_{\pm 1.2842}$ & \textbf{0.9492} & $_{\pm 0.1648}$ & \textbf{0.1867} & $_{\pm 0.0047}$ \\
        & \textit{Corrective} Diffusion & 0.9292 & $_{\pm 0.1019}$ & 0.1327 & $_{\pm 0.4484}$ & \textbf{6.0387} & $_{\pm 2.3348}$ & 0.8608 & $_{\pm 0.1742}$ & \underline{0.1883} & $_{\pm 0.0106}$ \\
        & SerpentFlow                   & 0.9375 & $_{\pm 0.0029}$ & -0.1257 & $_{\pm 0.0391}$ & 7.8810 & $_{\pm 0.6203}$ & 0.7899 & $_{\pm 0.0456}$ & 0.1988 & $_{\pm 0.0014}$ \\
        & Consistency                   & \textbf{0.8667} & $_{\pm 0.0051}$ & \underline{0.0749} & $_{\pm 0.1336}$ & 12.3931 & $_{\pm 1.7233}$ & \underline{1.0613} & $_{\pm 0.0574}$ & 0.1980 & $_{\pm 0.0094}$ \\
        \midrule
        \multirow{6}{*}{AVERAGE} 
        & Bilinear                      & 0.7895 & & -0.1408 & & 18.2486 & & \multicolumn{2}{c}{-} & \multicolumn{2}{c}{-} \\
        & UNet                          & 0.6499 & $_{\pm 0.0037}$ & -0.4928 & $_{\pm 0.0248}$ & 17.8444 & $_{\pm 0.3891}$ & \multicolumn{2}{c}{-} & \multicolumn{2}{c}{-} \\
        & Diffusion                     & \textbf{0.5289} & $_{\pm 0.0048}$ & \textbf{-0.0093} & $_{\pm 0.1699}$ & \underline{5.9118} & $_{\pm 0.9349}$ & \textbf{0.9351} & $_{\pm 0.0885}$ & \underline{0.1488} & $_{\pm 0.0024}$ \\
        & \textit{Corrective} Diffusion & 0.5446 & $_{\pm 0.0354}$ & 0.0260 & $_{\pm 0.1685}$ & \textbf{5.6255} & $_{\pm 1.2814}$ & 0.8242 & $_{\pm 0.0790}$ & \textbf{0.1480} & $_{\pm 0.0039}$ \\
        & SerpentFlow                   & 0.5748 & $_{\pm 0.0016}$ & \underline{-0.0122} & $_{\pm 0.0232}$ & 9.6271 & $_{\pm 0.6492}$ & \underline{0.8875} & $_{\pm 0.0403}$ & 0.1611 & $_{\pm 0.0012}$ \\
        & Consistency                   & \underline{0.5375} & $_{\pm 0.0067}$ & 0.1663 & $_{\pm 0.0755}$ & 10.1096 & $_{\pm 0.7404}$ & 1.1337 & $_{\pm 0.0356}$ & 0.1598 & $_{\pm 0.0049}$ \\
        \bottomrule            
    \end{tabular}%
    }
\end{table}

\clearpage
\subsubsection{Randomly-subsampled test datasets}
\label{sec:appendix-indomain-random}

\begin{figure}[!htbp]
    \centering
    \includegraphics[width=1.0\linewidth]{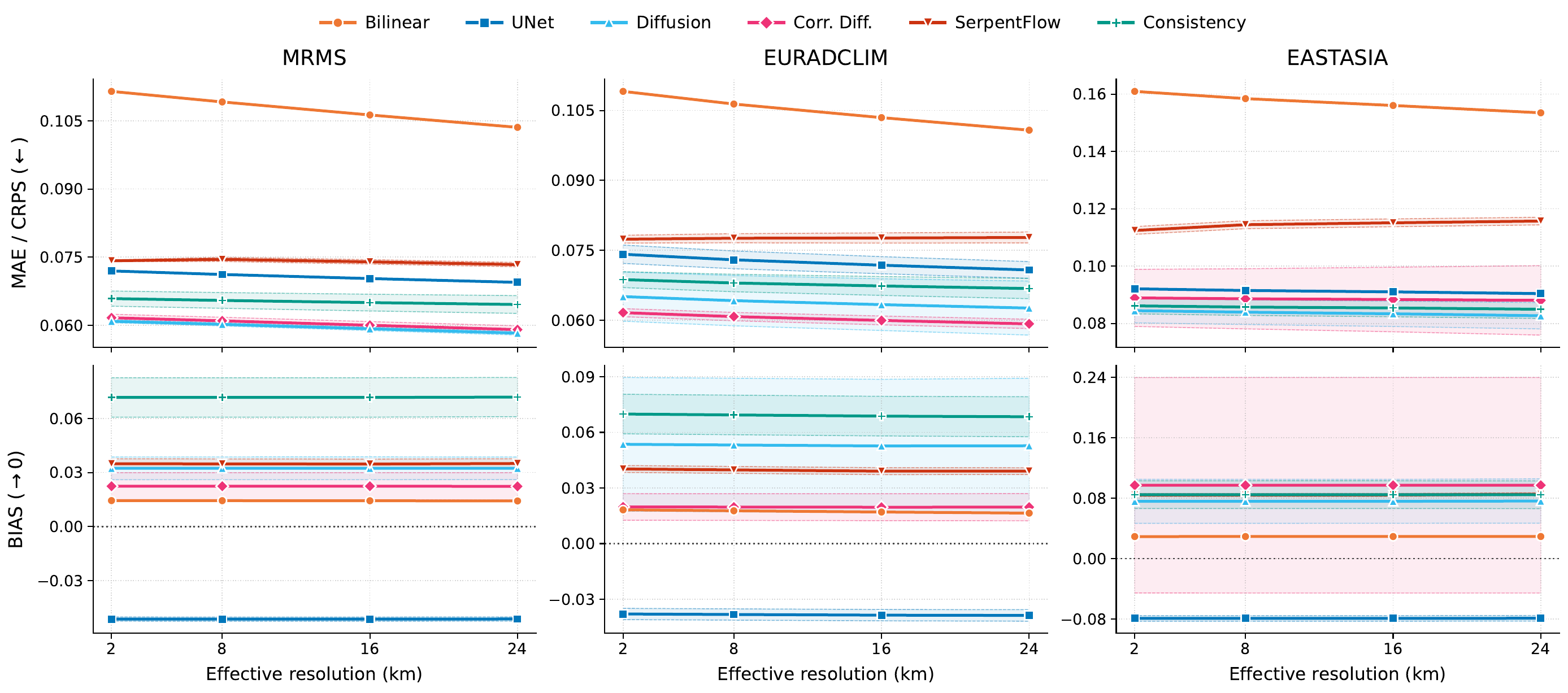}
    \caption{\textbf{Effect of spatial coarsening on pointwise error and bias (randomly-subsampled test sets).} Settings and results are similar to Figure \ref{fig:maecrpsbias-dsfs-importance}.}
    \label{fig:maecrpsbias-dsfs-random}
\end{figure}

\begin{figure}[!htbp]
  \centering
  \includegraphics[width=1.0\textwidth]{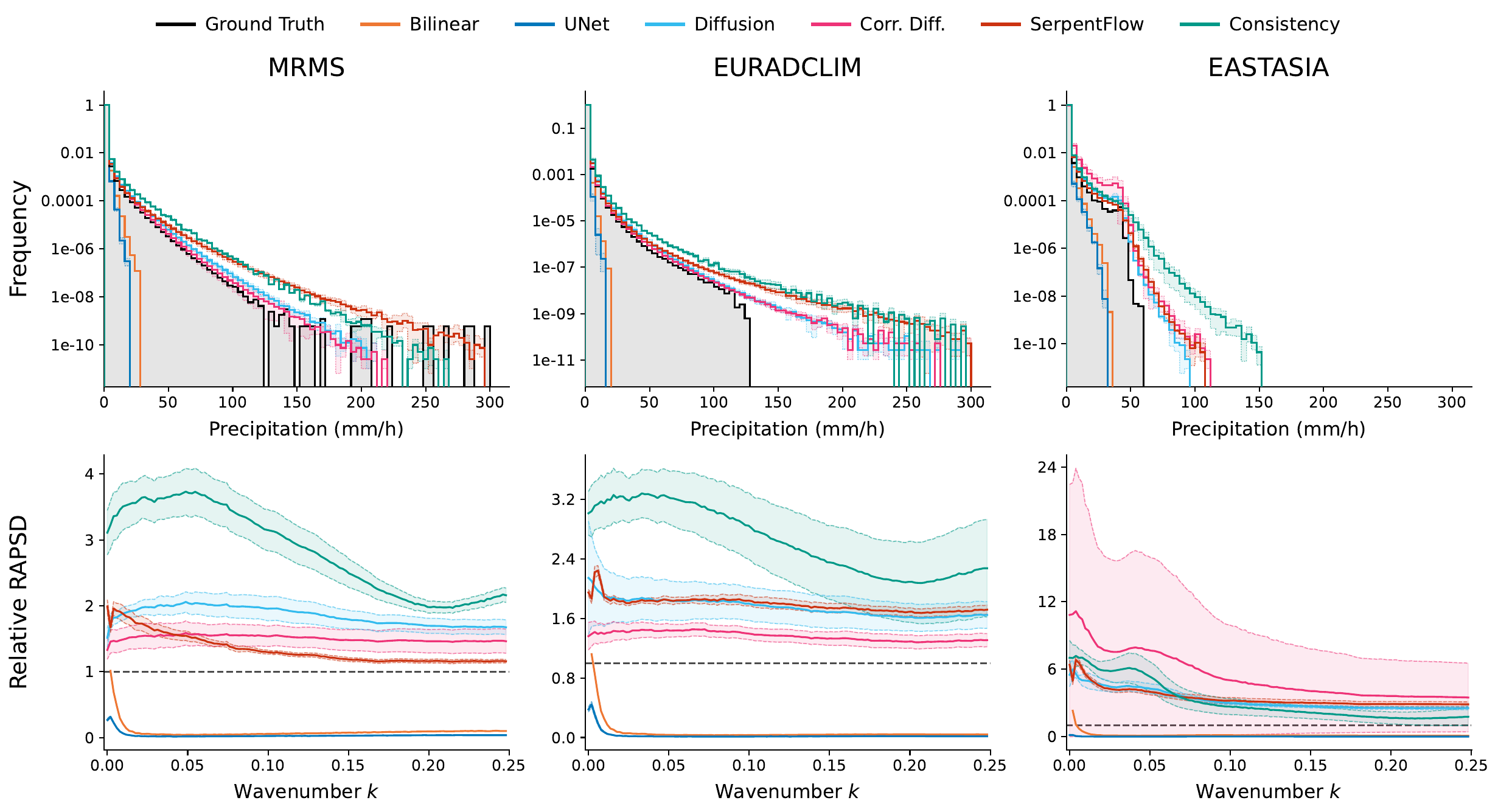}
  \caption{\textbf{In-domain intensity distribution and spectral fidelity (randomly-subsampled test sets).} Settings are similar to Figure \ref{fig:indomain-histograms-rapsd-importance}. Generative baselines show much larger RAPSD, with high variance on EA, compared to evaluation on importance-subsampled test datasets.}
  \label{fig:indomain-histograms-rapsd-random}
\end{figure}

\begin{figure}[!htbp]
    \centering
    \includegraphics[width=1.0\linewidth]{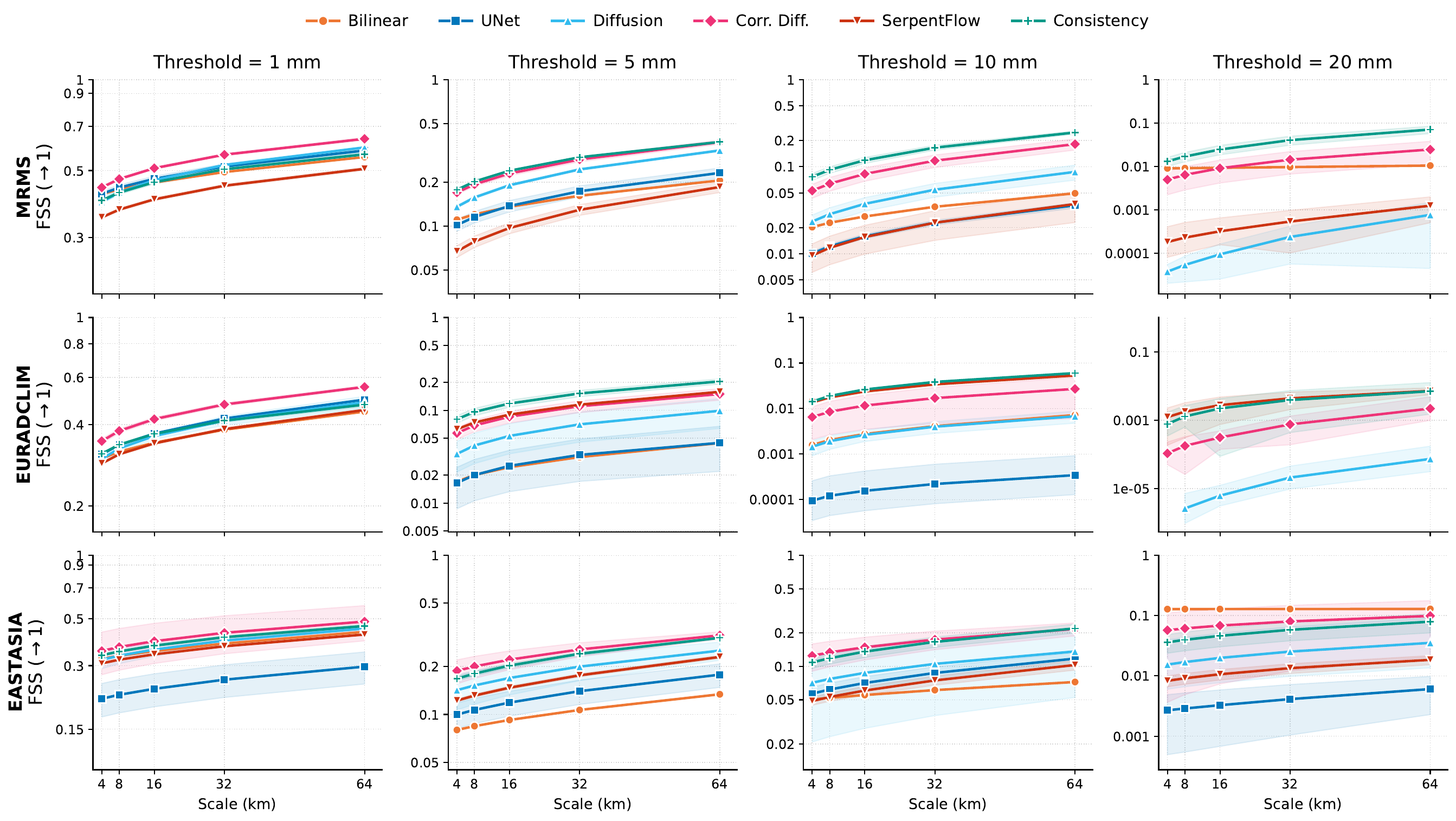}
    \caption{\textbf{Fractions Skill Score across precipitation thresholds (randomly-subsampled test sets).} Settings are similar to Figure \ref{fig:fss-importance}. Generative models maintain notable skill for higher intensity thresholds even under climatologically unweighted random sampling.}
    \label{fig:fss-random}
\end{figure}

\begin{figure}[!htbp]
    \centering
    \includegraphics[width=1.0\linewidth]{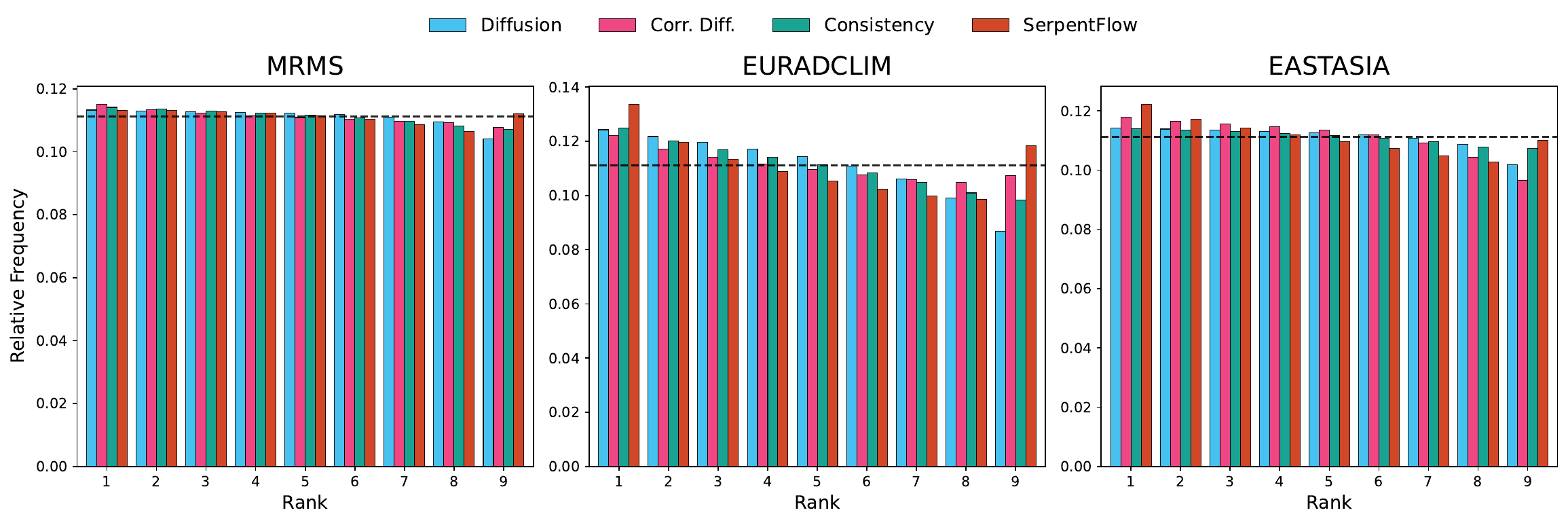}
    \caption{\textbf{Ensemble calibration via rank histograms (randomly-subsampled test sets).} Settings are similar to Figure \ref{fig:rank-histos-importance}. Ensemble calibration patterns remain consistent with the importance-subsampled results.}
    \label{fig:rank-histo-random}
\end{figure}

To evaluate model performance under unweighted climatological conditions dominated by dry and light precipitation events, we report in-domain results on the randomly-subsampled test sets. Table~\ref{tab:indomain-results-random} lists numerical scores across all baselines. Figure~\ref{fig:maecrpsbias-dsfs-random} illustrates the sensitivity of MAE/CRPS and bias across spatial aggregation scales. Figure~\ref{fig:indomain-histograms-rapsd-random} displays intensity histograms and RAPSD curves. Figure~\ref{fig:fss-random} details the scale-dependent Fractions Skill Score across intensity thresholds, and Figure~\ref{fig:rank-histo-random} presents rank histograms evaluating ensemble calibration under random sampling.

\begin{table}[!htbp]
    \caption{\textbf{In-domain performance evaluation on randomly-subsampled test datasets.} Details regarding metrics and formatting are identical to Table~\ref{tab:indomain-results-importance}. Probabilistic metrics are evaluated exclusively for generative models. The best and second-best results are \textbf{bolded} and \underline{underlined}, respectively. Subscripts denote standard deviations.}
    \label{tab:indomain-results-random}
    \centering
    \resizebox{\textwidth}{!}{%
    \begin{tabular}{l l r@{\,\,}l r@{\,\,}l r@{\,\,}l r@{\,\,}l r@{\,\,}l} 
        \toprule
        & & \multicolumn{2}{c}{MAE / CRPS $(\downarrow)$} & \multicolumn{2}{c}{BIAS $(\rightarrow 0)$} & \multicolumn{2}{c}{LHD $(\downarrow)$} & \multicolumn{2}{c}{SSR $(\rightarrow 1)$} & \multicolumn{2}{c}{VARIOGRAM $(\downarrow)$} \\
        \midrule
        \multirow{6}{*}{MRMS} 
        & Bilinear                      & 0.1115 & & \textbf{0.0144} & & 18.7129 & & \multicolumn{2}{c}{-} & \multicolumn{2}{c}{-} \\            
        & UNet                          & 0.0719 & $_{\pm 0.0001}$ & -0.0513 & $_{\pm 0.0011}$ & 19.8704 & $_{\pm 0.1937}$ & \multicolumn{2}{c}{-} & \multicolumn{2}{c}{-} \\
        & Diffusion                     & \textbf{0.0609} & $_{\pm 0.0003}$ & 0.0323 & $_{\pm 0.0063}$ & \underline{5.0397} & $_{\pm 0.6527}$ & 1.2478 & $_{\pm 0.0158}$ & \underline{0.0314} & $_{\pm 0.0004}$ \\
        & \textit{Corrective} Diffusion & \underline{0.0616} & $_{\pm 0.0008}$ & \underline{0.0224} & $_{\pm 0.0075}$ & \textbf{3.6442} & $_{\pm 1.6567}$ & \textbf{1.0549} & $_{\pm 0.0788}$ & \textbf{0.0304} & $_{\pm 0.0006}$ \\
        & SerpentFlow                   & 0.0742 & $_{\pm 0.0002}$ & 0.0349 & $_{\pm 0.0028}$ & 9.2778 & $_{\pm 0.5459}$ & \underline{1.2290} & $_{\pm 0.0520}$ & 0.0340 & $_{\pm 0.0005}$ \\
        & Consistency                   & 0.0659 & $_{\pm 0.0017}$ & 0.0716 & $_{\pm 0.0109}$ & 7.3231 & $_{\pm 0.2379}$ & 1.4877 & $_{\pm 0.0318}$ & 0.0347 & $_{\pm 0.0008}$ \\
        \midrule
        \multirow{6}{*}{EURADCLIM} 
        & Bilinear                      & 0.1091 & & \textbf{0.0182} & & 18.7759 & & \multicolumn{2}{c}{-} & \multicolumn{2}{c}{-} \\
        & UNet                          & 0.0741 & $_{\pm 0.0020}$ & -0.0379 & $_{\pm 0.0030}$ & 19.9062 & $_{\pm 0.4600}$ & \multicolumn{2}{c}{-} & \multicolumn{2}{c}{-} \\
        & Diffusion                     & \underline{0.0650} & $_{\pm 0.0053}$ & 0.0535 & $_{\pm 0.0360}$ & \underline{5.3655} & $_{\pm 1.4171}$ & 1.2050 & $_{\pm 0.1036}$ & \underline{0.0308} & $_{\pm 0.0031}$ \\
        & \textit{Corrective} Diffusion & \textbf{0.0616} & $_{\pm 0.0009}$ & \underline{0.0198} & $_{\pm 0.0072}$ & \textbf{5.3510} & $_{\pm 0.1151}$ & \textbf{0.9797} & $_{\pm 0.0232}$ & \textbf{0.0275} & $_{\pm 0.0004}$ \\
        & SerpentFlow                   & 0.0774 & $_{\pm 0.0008}$ & 0.0403 & $_{\pm 0.0019}$ & 9.4164 & $_{\pm 0.8482}$ & \underline{0.9559} & $_{\pm 0.0145}$ & 0.0358 & $_{\pm 0.0008}$ \\
        & Consistency                   & 0.0687 & $_{\pm 0.0017}$ & 0.0699 & $_{\pm 0.0107}$ & 9.9300 & $_{\pm 0.6042}$ & 1.3335 & $_{\pm 0.0390}$ & 0.0325 & $_{\pm 0.0015}$ \\
        \midrule
        \multirow{6}{*}{EASTASIA} 
        & Bilinear                      & 0.1609 & & \textbf{0.0291} & & 14.2227 & & \multicolumn{2}{c}{-} & \multicolumn{2}{c}{-} \\
        & UNet                          & 0.0921 & $_{\pm 0.0002}$ & -0.0793 & $_{\pm 0.0037}$ & 16.0623 & $_{\pm 1.2856}$ & \multicolumn{2}{c}{-} & \multicolumn{2}{c}{-} \\
        & Diffusion                     & \textbf{0.0845} & $_{\pm 0.0042}$ & \underline{0.0758} & $_{\pm 0.0291}$ & \underline{7.7125} & $_{\pm 0.4733}$ & 1.3349 & $_{\pm 0.0504}$ & \textbf{0.0284} & $_{\pm 0.0013}$ \\
        & \textit{Corrective} Diffusion & 0.0889 & $_{\pm 0.0100}$ & 0.0970 & $_{\pm 0.1427}$ & \textbf{6.4958} & $_{\pm 3.0891}$ & \underline{1.1446} & $_{\pm 0.4056}$ & 0.0306 & $_{\pm 0.0094}$ \\
        & SerpentFlow                   & 0.1124 & $_{\pm 0.0013}$ & 0.0843 & $_{\pm 0.0019}$ & 8.9906 & $_{\pm 0.2691}$ & \textbf{1.0521} & $_{\pm 0.0231}$ & 0.0380 & $_{\pm 0.0011}$ \\
        & Consistency                   & \underline{0.0861} & $_{\pm 0.0028}$ & 0.0846 & $_{\pm 0.0184}$ & 12.7561 & $_{\pm 1.3035}$ & 1.3858 & $_{\pm 0.0477}$ & \underline{0.0299} & $_{\pm 0.0017}$ \\
        \midrule
        \multirow{6}{*}{AVERAGE} 
        & Bilinear                      & 0.1272 & & \textbf{0.0206} & & 17.2372 & & \multicolumn{2}{c}{-} & \multicolumn{2}{c}{-} \\
        & UNet                          & 0.0794 & $_{\pm 0.0007}$ & -0.0562 & $_{\pm 0.0026}$ & 18.6130 & $_{\pm 0.6464}$ & \multicolumn{2}{c}{-} & \multicolumn{2}{c}{-} \\
        & Diffusion                     & \textbf{0.0701} & $_{\pm 0.0033}$ & 0.0539 & $_{\pm 0.0238}$ & \underline{6.0392} & $_{\pm 0.8477}$ & 1.2626 & $_{\pm 0.0566}$ & \underline{0.0302} & $_{\pm 0.0016}$ \\
        & \textit{Corrective} Diffusion & \underline{0.0707} & $_{\pm 0.0039}$ & \underline{0.0464} & $_{\pm 0.0524}$ & \textbf{5.1637} & $_{\pm 1.6203}$ & \textbf{1.0598} & $_{\pm 0.1692}$ & \textbf{0.0295} & $_{\pm 0.0035}$ \\
        & SerpentFlow                   & 0.0880 & $_{\pm 0.0008}$ & 0.0531 & $_{\pm 0.0022}$ & 9.2283 & $_{\pm 0.5544}$ & \underline{1.0790} & $_{\pm 0.0299}$ & 0.0360 & $_{\pm 0.0008}$ \\
        & Consistency                   & 0.0736 & $_{\pm 0.0020}$ & 0.0754 & $_{\pm 0.0133}$ & 10.0031 & $_{\pm 0.7152}$ & 1.4023 & $_{\pm 0.0395}$ & 0.0324 & $_{\pm 0.0013}$ \\
        \bottomrule            
    \end{tabular}%
    }
\end{table}

\clearpage
\subsection{Geographical generalization}
\label{sec:appendix-geographical-generalization}

\begin{figure}[!htbp]
    \centering
    \includegraphics[width=1.0\linewidth]{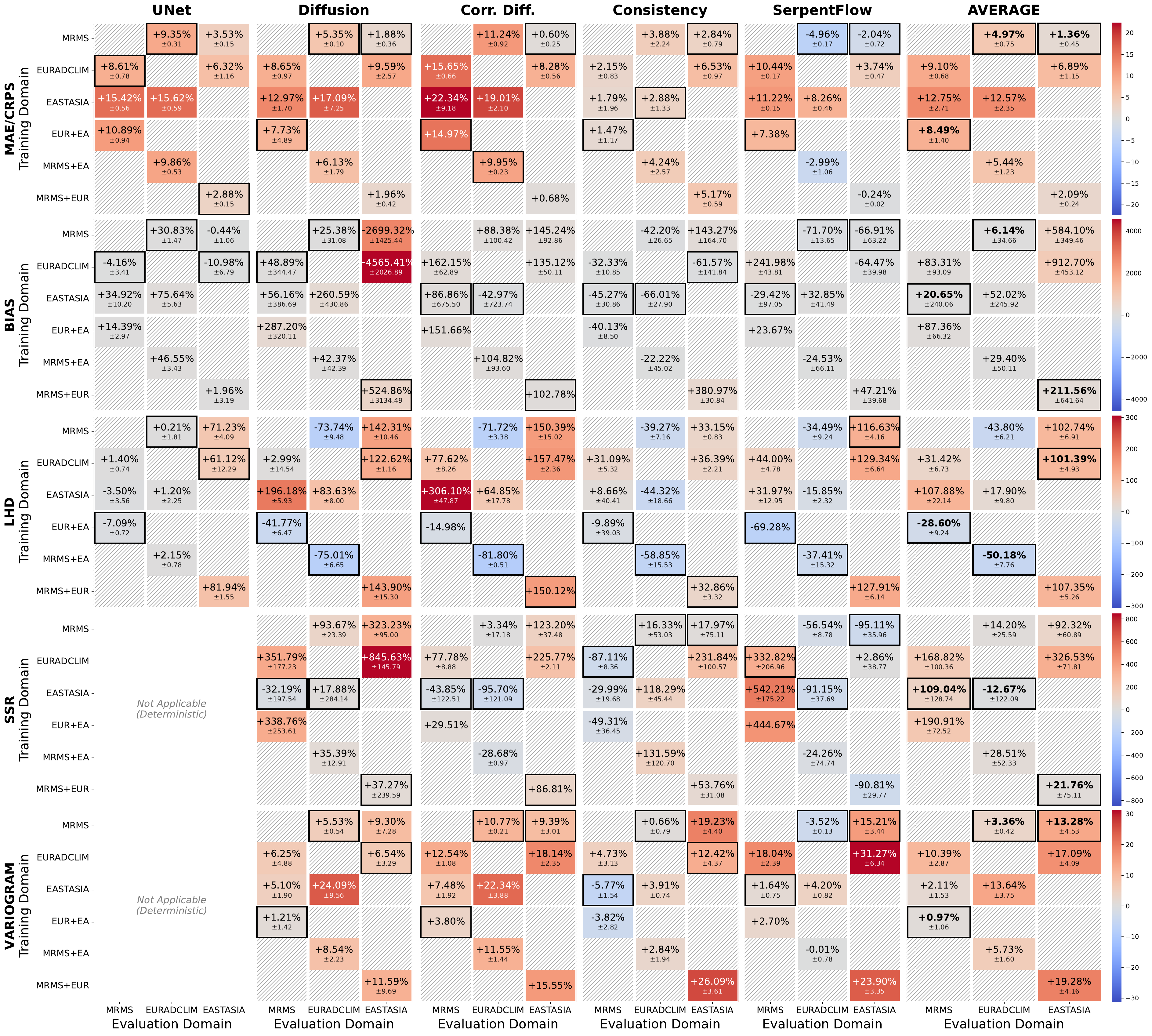}
    \caption{\textbf{Out-of-domain generalization relative degradation on importance-subsampled test sets.} Relative degradation ($\%$) with respect to in-domain training across five evaluation metrics (MAE/CRPS, BIAS, LHD, SSR, and VARIOGRAM). Best-performing out-of-domain training domains per column are boxed in black. Scores are averaged over three random seeds, except for configurations with runs still in progress at submission, which are averaged over the seeds available. Full results will be updated during the rebuttal period.}
    \label{fig:appendix-generalization-importance}
\end{figure}

\begin{figure}[!htbp]
    \centering
    \includegraphics[width=1.0\linewidth]{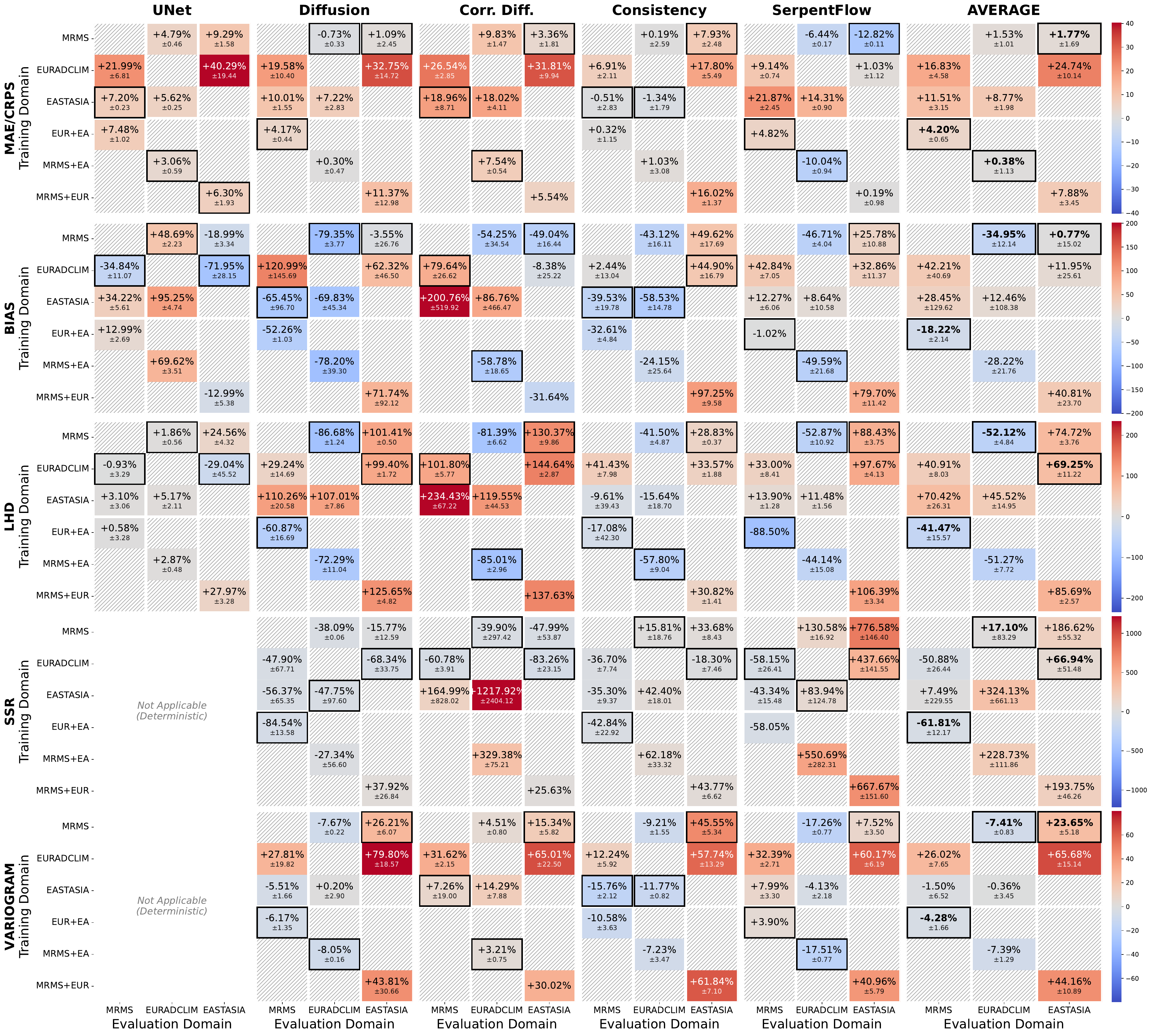}
    \caption{\textbf{Out-of-domain generalization relative degradation on randomly-subsampled test sets.} Relative degradation ($\%$) with respect to in-domain training across five evaluation metrics (MAE/CRPS, BIAS, LHD, SSR, and VARIOGRAM). Scores are averaged over three random seeds, except for configurations with runs still in progress at submission, which are averaged over the seeds available. Full results will be updated during the rebuttal period.}
    \label{fig:appendix-generalization-random}
\end{figure}

\begin{figure}[!htbp]
    \centering
    \includegraphics[width=1.0\linewidth]{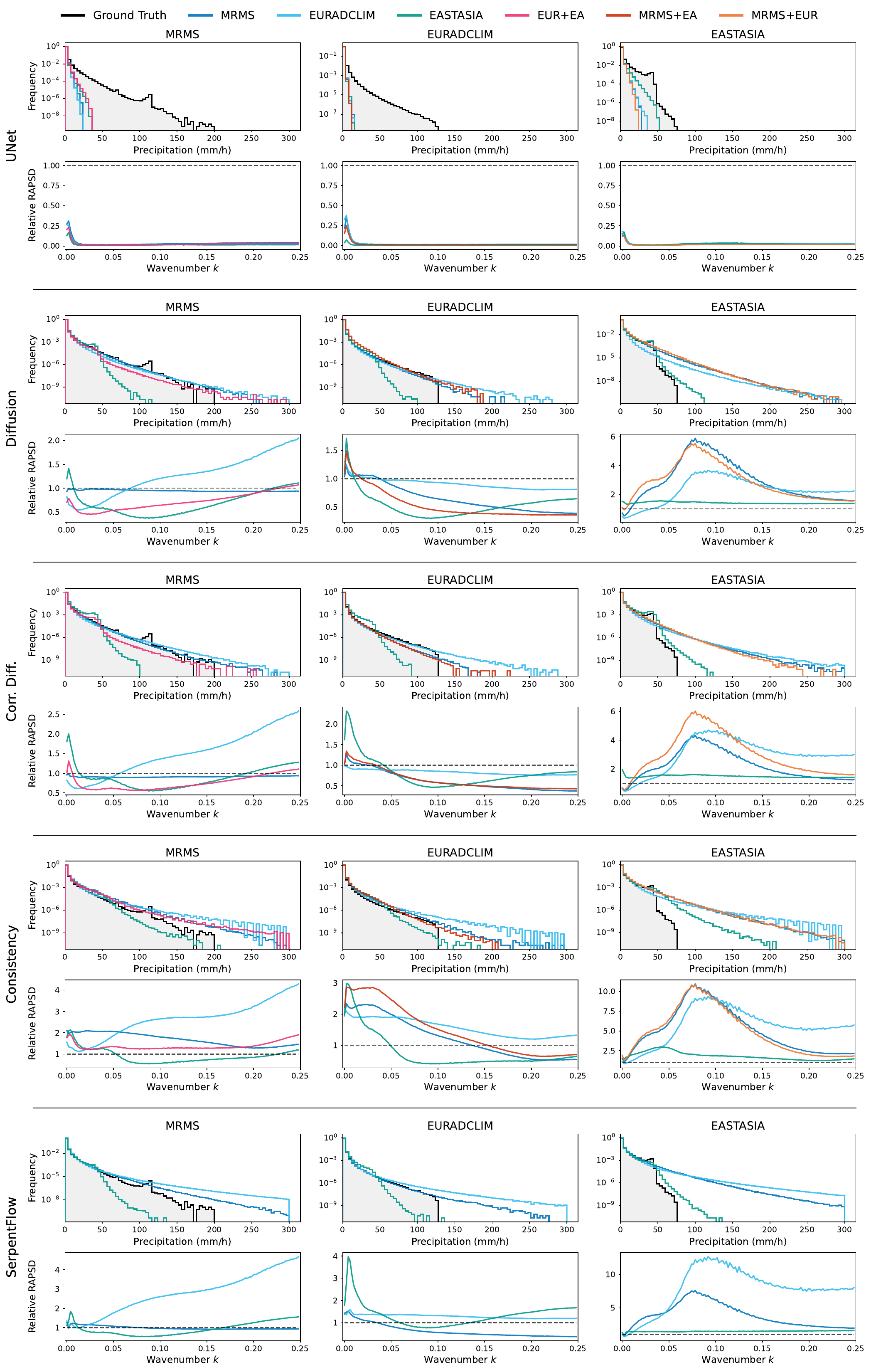}
    \caption{\textbf{Intensity histograms and relative RAPSD under geographical generalization (importance-subsampled test sets).} Line colors distinguish the different training domains (MRMS, EURADCLIM, EASTASIA, EUR+EA, MRMS+EA, MRMS+EUR) against ground truth (black).}
    \label{fig:generalization-histos-rapsd}
\end{figure}

This section provides extended results on geographical generalization. Figures~\ref{fig:appendix-generalization-importance} and~\ref{fig:appendix-generalization-random} present relative degradation heatmaps across all five evaluation metrics for importance-subsampled and randomly-subsampled test sets, respectively. Figure~\ref{fig:generalization-histos-rapsd} details the resulting precipitation intensity histograms and relative power spectral density curves under geographical shift for importance-subsampled test sets.

\end{document}